\documentclass{ieeeaccess}
\usepackage{cite}
\usepackage{amsmath,amssymb,amsfonts}
\usepackage{algorithmic}
\usepackage{graphicx}
\usepackage{algorithm}
\usepackage{algorithmic}
\usepackage{textcomp}
\usepackage{comment}
\newcommand{\etal}{\textit{et al}.}

\usepackage{multirow}
\usepackage{booktabs}
\usepackage[table,dvipsnames]{xcolor}
\usepackage{color,soul}

\def\BibTeX{{\rm B\kern-.05em{\sc i\kern-.025em b}\kern-.08em
    T\kern-.1667em\lower.7ex\hbox{E}\kern-.125emX}}
\begin{document}
\history{Date of publication xxxx 00, 0000, date of current version xxxx 00, 0000.}
\doi{10.1109/ACCESS.2017.DOI}

\title{A Two-Block RNN-based Trajectory Prediction from Incomplete Trajectory}
\author{\uppercase{Ryo Fujii\authorrefmark{1}, Jayakorn Vongkulbhisal\authorrefmark{2}, Ryo Hachiuma\authorrefmark{1}\IEEEmembership{Student Member, IEEE}, and Hideo Saito\authorrefmark{1}\IEEEmembership{Senior Member, IEEE}}}
\address[1]{Department of Science and Technology, Keio University, Yokohama, Kanagawa 223-8522, Japan}
\address[2]{IBM Research, Chuo-ku, Tokyo 103-8510, Japan}
\tfootnote{This work was supported by JST-Mirai Program Grant Number JPMJMI19B2, Japan.}

\markboth
{Author \headeretal: Preparation of Papers for IEEE TRANSACTIONS and JOURNALS}
{Author \headeretal: Preparation of Papers for IEEE TRANSACTIONS and JOURNALS}

\corresp{Corresponding author: Ryo Fujii (e-mail:ryo.fujii0112@keio.jp).}

\begin{abstract}
Trajectory prediction has gained great attention and significant progress has been made in recent years. However, most works rely on a key assumption that each video is successfully preprocessed by detection and tracking algorithms and the complete observed trajectory is always available. However, in complex real-world environments, we often encounter miss-detection of target agents (e.g., pedestrian, vehicles) caused by the bad image conditions, such as the occlusion by other agents. In this paper, we address the problem of trajectory prediction from incomplete observed trajectory due to miss-detection, where the observed trajectory includes several missing data points. We introduce a two-block RNN model that approximates the inference steps of the Bayesian filtering framework and seeks the optimal estimation of the hidden state when miss-detection occurs. The model uses two RNNs depending on the detection result. One RNN approximates the inference step of the Bayesian filter with the new measurement when the detection succeeds, while the other does the approximation when the detection fails. Our experiments show that the proposed model improves the prediction accuracy compared to the three baseline imputation methods on publicly available datasets: ETH and UCY ($9\%$ and $7\%$ improvement on the ADE and FDE metrics). We also show that our proposed method can achieve better prediction compared to the baselines when there is no miss-detection.
\end{abstract}


\titlepgskip=-15pt

\maketitle

\section{Introduction}
\label{sec:introduction}
Predicting future trajectory from video data is an indispensable technology for developing navigation systems that can be used in several scenarios, such as self-driving vehicles, social robots, and navigation systems for blind people. High-quality predictions guide the user to the appropriate path and avoid dangerous situations (e.g., collision).  The most common setting of trajectory prediction is the surveillance setting from a fixed camera, where the position of the agent (e.g., pedestrian, vehicles) is often treated as a single point \cite{Alahi_2016_CVPR, Gupta_2018_CVPR}. 

\begin{figure}[tb]
 \centering
 \includegraphics[width = \linewidth]{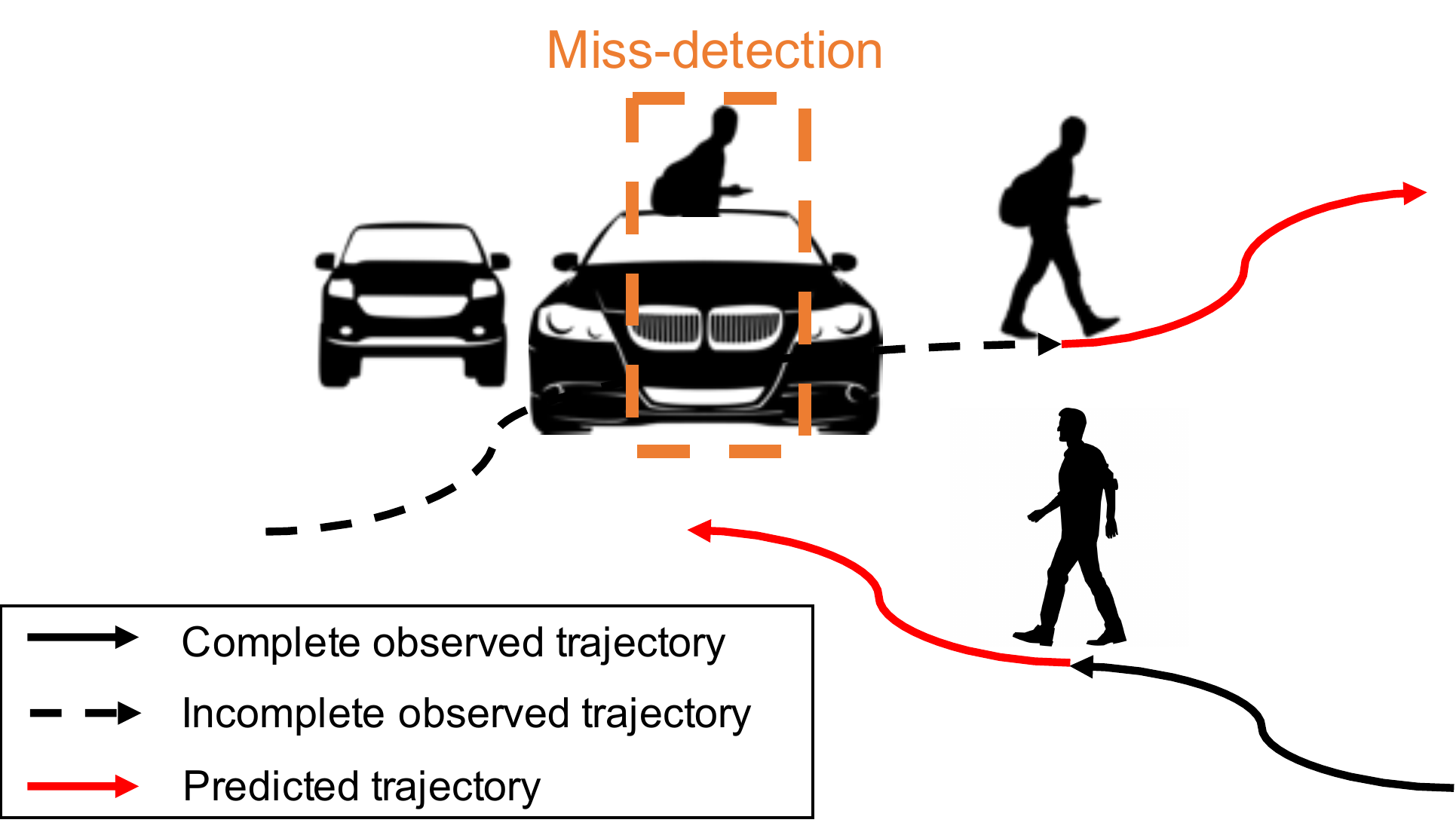}
 \caption{Our goal is to predict the future trajectory of agents from incomplete trajectory because of miss-detection. We develop a model which does not affect the trajectory prediction result from complete trajectory.}
 \label{fig:1}
\end{figure}

\begin{figure*}[tb]
\begin{tabular}{ccc}
\begin{minipage}{0.3\hsize}
    \begin{center}
        \includegraphics[clip, width=\hsize]{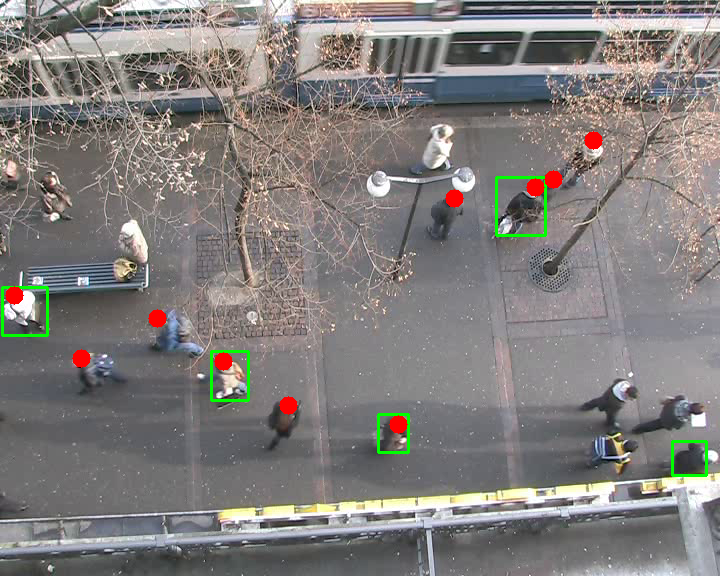} 
    \end{center}
\end{minipage}
&
\begin{minipage}{0.3\hsize}
    \begin{center}
        \includegraphics[clip, width=\hsize]{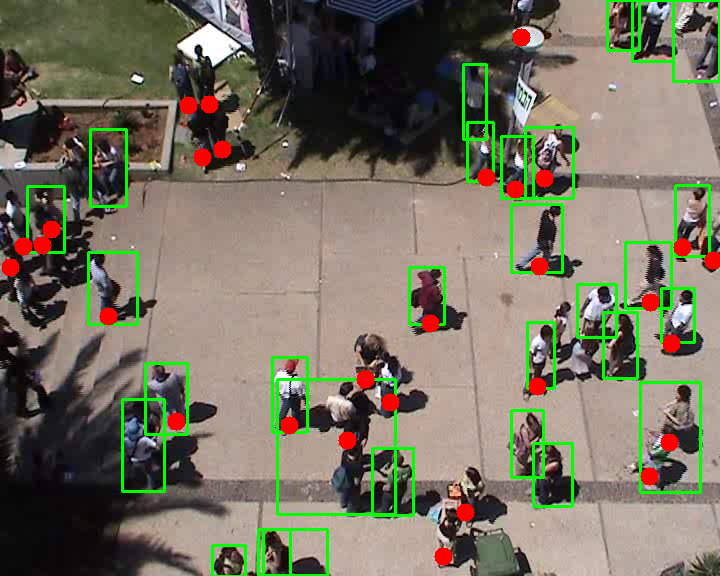}
    \end{center}
\end{minipage}
&
\begin{minipage}{0.3\hsize}
    \begin{center}
        \includegraphics[clip, width=\hsize]{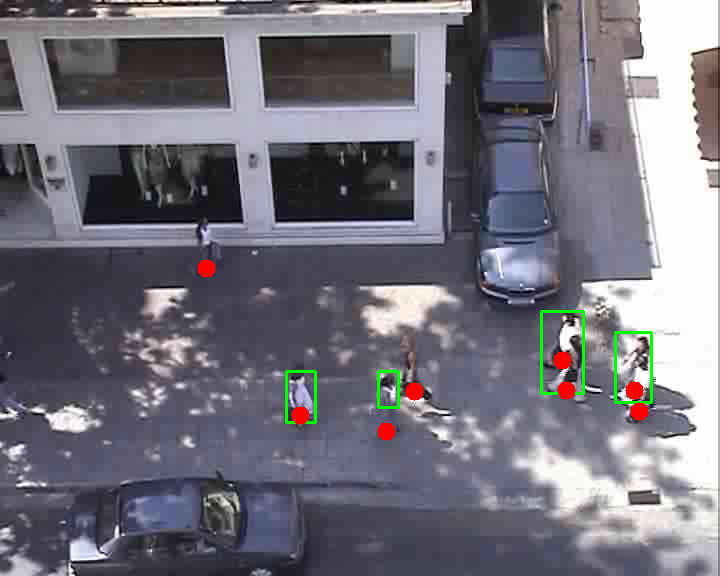} 
    \end{center}
\end{minipage}
\end{tabular}
\caption{Examples of miss-detection annotated by Faster R-CNN \cite{Ren_2016_TPAMI} pretrained with
MS-COCO dataset \cite{Lin_2014_ECCV} from ETH \cite{Pellegrini_2010_ECCV} and UCY \cite{Leal_2014_CVPR} datasets. The red points indicate the ground-truth positions attached to each dataset and green boxes indicate the predictions. Existing works on trajectory prediction often ignore miss-detected agents.}
 \label{fig:2}
\end{figure*}

Many approaches to trajectory prediction forecast the future state of the target agent conditioned on the history of past states \cite{Kooij_2019_IJCV, Karasev_2016_ICRA, Pellegrini_2009_ICCV, Kitani_ECCV_2012}. The state of the agent is often represented as its position. To obtain the history of a position of the agents, we need to detect where the agents are in the current and past time steps (e.g., object detection) and to establish object correspondences between time steps (e.g., object tracking). Therefore, the trajectory prediction task is a downstream task of object detection and object tracking, and greatly depends on the performance of the upstream tasks. Many trajectory prediction works rely on a key assumption that each sequence is successfully preprocessed by object detection and tracking algorithms,and that they can always access the complete trajectory for trajectory prediction. In other words, the detection algorithm has to successfully detect all of the agents in the scene, and the tracking algorithm has to successfully track all of the detected agents. In details, most trajectory prediction research from bird's-eye view images assumes that all positions of all pedestrians can be obtained during $6.4$ or $8$ seconds: observing the trajectory for $3.2$ seconds and predicting the future trajectory for $3.2$ or $4.8$ seconds \cite{Alahi_2016_CVPR, Gupta_2018_CVPR}. However, this is not feasible for real-world applications. There often emerge cases wherein an object is miss-detected due to bad image condition such as motion blurs, illumination changes, occlusion by other agents, and cluttered backgrounds \cite{Hosoya_2020_WACV}, as shown in Fig. \ref{fig:2}. When the target agent is not detected in several frames during the observation time, the future trajectory cannot be predicted against incomplete observed trajectory (Fig. \ref{fig:1}).

The common solution for dealing with incomplete observed trajectory is to ignore these cases from the dataset as outliers. The agent who disappears even in one frame in a sequence is excluded from the dataset to handle incomplete data  \cite{Alahi_2016_CVPR, Gupta_2018_CVPR, Styles_2020_WACV}. However, in real-world situations, an intelligent system must be able to continuously predict the future trajectory of all agents. This exclusion could cause the system to ignore the interaction between the excluded target agent and other agents. Furthermore, the system cannot consider the possibility of collision with the excluded agents and it could cause dangerous accidents. To avoid the exclusion of an undetected agent,
we could attempt to impute the missing state of the agent. One possible solution is to impute a previously observed state when the miss-detection occurs\cite{Yao_2019_IROS, Malla_2020_CVPR}. This can prevent the data exclusion of the agent whose state is not available. However, this can cause the model to interpret the data as the person keeping the previous state. Thus, compensating the states with an assumed value can lead to high error. In this paper, we investigate the problem of trajectory prediction from incomplete observations particularly due to miss-detection.

Bayesian filter-based methods \cite{Candy_2009}  (particularly, Kalman filter \cite{kalman1960}) are among the traditional approaches for trajectory prediction \cite{Schneider_2013_PR} and they are often used as baseline methods for comparison \cite{Alahi_2016_CVPR}. Bayesian filters recursively update the posterior distribution of predictions with the arrival of new data. The filtering process can be described as a cycle of two steps, the prediction and the update step. Due to simple structure, they do not often perform well on long-term prediction \cite{Barth_2008_IV}. With the ability to learn and produce long sequences, Recurrent Neural Networks (RNNs), and in particular the Long Short-Term Memory (LSTM) networks \cite{Hochreiter_1997_NC}, have recently become a widely popular modeling approach for predicting human motion \cite{Alahi_2016_CVPR, Bartoli_2018_ICPR, Vemula_2018_ICRA, Bisagno_2019_ECCV, Sadeghian_2019_CVPR, Saleh_2018_IV,Sun_2018_ICRA,  Pfeiffer_2018_ICRA, Shi_2019_SENSORS, Zhang_2019_CVPR}. Recently, the connection between Bayesian filters and RNNs has been studied, and the observation that Bayesian filters are a special type of RNNs has been proposed \cite{Gu_2017_CVPR}. Inspired by the connection, we hypothesized that using the assumed state (e.g., the last observed state) for the missing time step causes the RNNs to fail to update their hidden state.

Using this intuition, to avoid wrong updates of the hidden state, we propose a simple two-block modification in which we add a new RNN block for the missing time step. We evaluate our two-block mechanism on an existing trajectory prediction model from a bird's-eye view against three baseline methods and show noticeable improvement to trajectory prediction from incomplete trajectories. Our proposed method does not affect trajectory prediction results from complete trajectories, even if the model is trained with incomplete data. Our contributions are as follows:


\begin{itemize}
  \item We propose a two-block RNN that learns the inference step of Bayesian filters for trajectory prediction from incomplete observed trajectory due to miss-detection.
  \item We show that our model can outperform three baselines and that it performs better compared to the best baseline by $12\%$ (ADE) and $4\%$ (FDE) on ETH \cite{Pellegrini_2010_ECCV} and UCY \cite{Leal_2014_CVPR}. Our model trained by incomplete trajectory does not affect the trajectory prediction result from the complete data.
\end{itemize}

 \begin{table*}
 
 \caption{A survey of approaches for handling missing entries in RNN-based time-series prediction tasks. See Sec.~{\ref{sec:ReviewMissing}} for details.
 }
\label{table:1}
\begin{center}

\label{table:survey}
\rowcolors{6}{gray!10}{white}
\begin{tabular}{cccccccc}
\toprule
\multirow{4}{*}{Approach} & \multicolumn{6}{c}{Method to handle missing entries} & \multirow{4}{*}{Target application}
\\
\cmidrule(l{2pt}r{3pt}){2-7} & \multirow{2.5}{*}{Drop data} & \multicolumn{3}{c}{Impute data for the encoding stage} & \multicolumn{2}{c}{Use modified RNNs} &
\\
\cmidrule(l{3pt}r{3pt}){3-5}\cmidrule(l{3pt}r{3pt}){6-7} & & Last Filling & Zero Filling & Learning-based & Internal & External
\\
\specialrule{.4pt}{2pt}{0pt}
Alahi \etal\cite{Alahi_2016_CVPR} & \checkmark & & & & & & Trajectory prediction
\\
Gupta \etal\cite{Gupta_2018_CVPR} & \checkmark & & & & & & Trajectory prediction
\\
Styles \etal\cite{Styles_2020_WACV} & \checkmark & & & & & & Trajectory prediction
\\
Yao \etal\cite{Yao_2019_IROS} & & \checkmark & & & & & 
Traffic accident detection
\\
Malla \etal\cite{Malla_2020_CVPR} & & \checkmark & & & & & Trajectory prediction
\\
Lipton \etal\cite{Lipton_2016_MLHC} & & \checkmark & \checkmark & &  & & Medical data analysis
\\
Kim \etal\cite{Kim_2017_BigComp} & & & & \checkmark & & & Medical data analysis
\\
Che \etal\cite{Che_2018_NATURE} & & \checkmark & & & \checkmark & & Medical data analysis
\\
Cao \etal\cite{Cao_2018_NEURIPS}& & & & \checkmark & \checkmark & & General
\\
Tian \etal\cite{Tian_2018_Neurocomputing}& & & & \checkmark & \checkmark & & Traffic flow prediction
\\
Kim \etal\cite{Kim_2018_IJCAI}& & & & \checkmark  & \checkmark & & Medical data analysis
\\
Luo \etal\cite{Luo_2018_NEURIPS}& & & & \checkmark & \checkmark & & General
\\
Luo \etal\cite{Luo_2019_IJCAI}& & & & \checkmark & \checkmark & & General
\\
Ours & & & & & & \checkmark & Trajectory prediction
\\
\specialrule{.8pt}{0pt}{2pt}
\end{tabular}

\end{center}

\end{table*}

\section{Related Work}
The problem of trajectory prediction has received significant attention in recent years across various applications, such as self-driving vehicles, service robots, and advanced surveillance systems. A large body of research has addressed this problem. Many approaches define an explicit dynamical model based on Newton’s laws of motion and use them as building blocks of a Bayesian filter (particularly, Kalman filter) \cite{Elnagar_2001_ICRA, Karasev_2016_ICRA, Barth_2008_IV, Batz_2009_IV}. Many other works have investigated how to incorporate the concept of a rational agent when modeling human motions. Recently, a number of works explore approaches to approximate motion dynamics from training data using deep neural networks \cite{Alahi_2016_CVPR, Bartoli_2018_ICPR, Vemula_2018_ICRA, Sadeghian_2019_CVPR, Saleh_2018_IV,Sun_2018_ICRA, Bisagno_2019_ECCV, Pfeiffer_2018_ICRA,Zhang_2019_CVPR,Shi_2019_SENSORS, Gupta_2018_CVPR, Rhinehart_ICCV_2019, Zhao_2019_CVPR, Huang_2019_ICCV, Kosaraju_2019_Neurips, Ivanovic_ICCV_2019}. In particular, Recurrent Neural Networks (RNNs) have recently become widely popular for modeling the dynamics approach \cite{Alahi_2016_CVPR, Bartoli_2018_ICPR, Vemula_2018_ICRA, Sadeghian_2019_CVPR, Saleh_2018_IV,Sun_2018_ICRA, Bisagno_2019_ECCV, Pfeiffer_2018_ICRA,Zhang_2019_CVPR,Shi_2019_SENSORS}. These sequential data-driven approaches assume $N^{th}$-order Markov models in which a limited state (e.g., position, velocity) history of $N$ time steps is a sufficient representation of the entire state history. In this section, we only review RNN-based trajectory prediction literature. For a more extensive review, we refer the reader to the article in \cite{Andrey_2019_CoRR}.

\subsection{RNNs for trajectory prediction}
Trajectory prediction has been studied extensively in surveillance settings from bird's-eye views. Many studies have developed models to account for agent social interactions and social convention, in addition to scene semantics that may affect the trajectory. Social-LSTM \cite{Alahi_2016_CVPR} is a pioneer work to model pedestrian trajectories as well as their interactions in continuous space. It introduces the social pooling layer, which allows the LSTMs to share the hidden states of the agents that are nearby. Many methods \cite{Gupta_2018_CVPR, Sadeghian_2019_CVPR, Liang_2019_CVPR_Workshops, Huang_2019_ICCV} follow the problem formulation used in \cite{Alahi_2016_CVPR}, including the assumption that each scene is first processed to obtain the spatial coordinates of all people at different time instants. Recently, to achieve multi-modality in the prediction output, many works combine RNN and Generative Adversarial Networks (GANs) \cite{Gupta_2018_CVPR, Sadeghian_2019_CVPR, Kosaraju_2019_Neurips}. Gupta \etal \cite{Gupta_2018_CVPR} proposed Social GAN with a new pooling mechanism that does take into account social interactions between all people in the scene and a variety loss that encourages the network to produce diverse predictions. Sophie \cite{Sadeghian_2019_CVPR} proposed an LSTM-based GAN with two attention mechanisms. Kosaraju \etal \cite{Kosaraju_2019_Neurips} proposed Social-BiGAT, which utilizes GAN with graph attention networks (GAT) that captures the social interaction.

\subsection{Handling missing data for RNN}
\label{sec:ReviewMissing}
In order to use RNNs for time-series prediction tasks, the observed data must be encoded by the RNNs before the prediction can be made.
However, missing entries in the observed data are unsuitable for encoding.
Many approaches have been developed to address this issue (see Table~{\ref{table:survey}}). 
If the task involves multiple time series (e.g., for trajectory prediction), a simple approach is to drop the missing data and perform the prediction using only the observed data~\cite{Alahi_2016_CVPR, Gupta_2018_CVPR, Styles_2020_WACV}. 
However, this strategy leads to loss of information and may not work if the missing rate is high or when there are individual time series to process (e.g., in medical applications).
Another strategy is to impute the missing data with some default values and to perform the prediction over the imputed data. 
For example, the last observed values~\cite{Yao_2019_IROS, Malla_2020_CVPR, Lipton_2016_MLHC,Che_2018_NATURE} or zero~\cite{Lipton_2016_MLHC} can be used to fill in the values of missing entries. 
Still, the filled values may cause the system to learn from them as if they are observed values, which could lead to lower performance.
More recent works propose to use learning-based techniques to perform imputation, e.g., using RNNs\cite{Kim_2017_BigComp,Cao_2018_NEURIPS,Tian_2018_Neurocomputing,Kim_2018_IJCAI} or GANs~\cite{Luo_2018_NEURIPS, Luo_2019_IJCAI} to estimate the missing entries, and they generally also propose to modify the internal structure of the RNNs, for example, to include a decay mechanism~\cite{Che_2018_NATURE,Tian_2018_Neurocomputing,Kim_2018_IJCAI} that puts different weights on data from different time steps. However, it may not be straightforward to apply the modification of one architecture to different architectures as this may involve different formulations or complicated implementations.

Unlike previous works that impute missing entries or modify internal structure of RNNs, in this work, we first look at the relation between Bayesian filter and RNNs, then derive an algorithm from the relation when there is a missing observation. This results in a simple-to-implement method that does not modify any internal structure of RNNs and also does not require an explicit form of imputation. Instead, our approach modifies RNNs externally by using two RNNs instead of one. This external modification allows our approach to be used with any RNN models.

\subsection{RNNs and Bayesian Filter}
Recently, the relationship between Bayesian filter and RNNs has been discussed. Gu \etal \cite{Gu_2017_CVPR} show Bayesian filter is a special type of RNN  and propose an RNN-based model for a face landmark localization task in videos. Lim \etal \cite{Lim_2020_IJCNN} introduced Recurrent Neural Filter (RNF), which aligns network modules with the inference steps of the Bayesian filter. RNF uses separate neural network components to directly model the Bayesian filtering steps. In this paper, we align RNN-based trajectory prediction models with the Bayesian filtering steps and explore the architecture that is suitable for trajectory prediction from incomplete observed trajectory. 

\begin{figure}[tb]
 \centering
 \includegraphics[width = \linewidth]{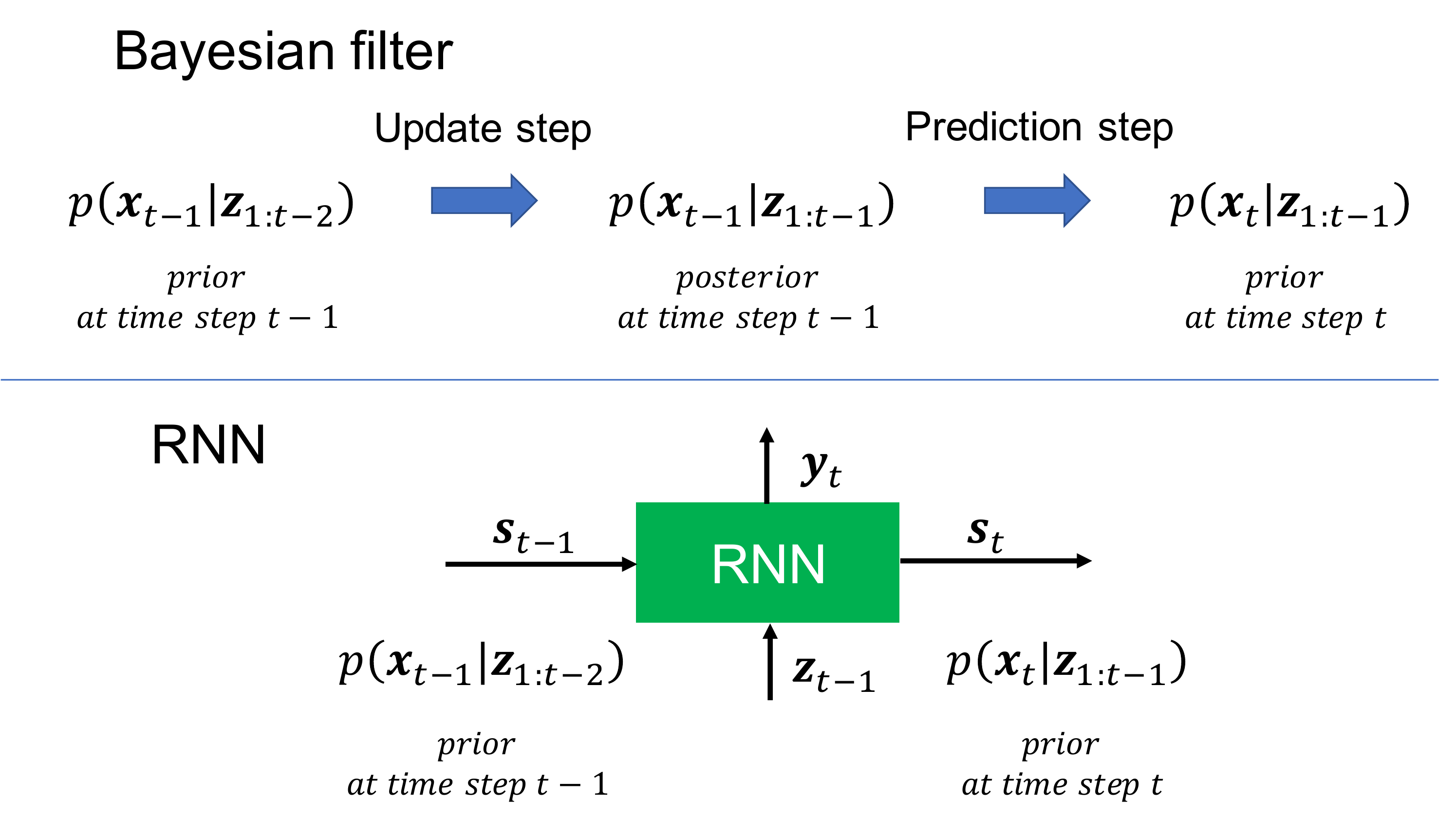}
 \caption{Relationship between Bayesian filter and RNNs. Given a sequence of measurement ${\bf z_t}$, Bayesian filter recursively estimate the optimal hidden states ${\bf x_t}$ through two steps: prediction and update step. Similarly, given a sequence of input ${\bf z_t}$, RNN updates the hidden state ${\bf s_t}$ every time step.}
 \label{fig:3}
\end{figure}

\section{Preliminaries}
In this section, we briefly review the Bayesian filter \cite{Candy_2009} and its connection with RNNs. The Bayesian filter has been used in a wide range of applications, including target tracking \cite{Haug_2012}, robotics \cite{Barfoot_2017}, and economics\cite{Gurnain_2006}. The goal of the Bayesian filter is to find the probable posterior distribution of the hidden state ${\bf x_t}$ at time instant $t$, given all of the measurements ${\bf z_{1:t}}$, which is characterized by $p({\bf x_t}|{\bf z_{1:t})}$. We use ${\bf z_{1:t}}$ to denote the sequence of measurements up to time instant $t$. Suppose there is a dynamical system represented by the following equations:
\begin{align} 
{\bf x_t}=&f_t({\bf x_{t-1}}, {\bf u_t}),  \\ 
{\bf z_t}=&h_t({\bf x_t}, {\bf v_t}),
\end{align}
where ${\bf u_t}$ and ${\bf v_t}$ are the system process noise and observation noise, respectively. Both are assumed to have known probability distributions. The functions $f$ and $h$ are the state transition function and the observation function, which can be represented in a probabilistic form as $p({\bf x_t}|{\bf x_{t-1}})$ and $p({\bf z_t}|{\bf x_t})$, respectively. When the state transition and the observation functions are linear and the process and measurement noises are Gaussian, the Bayesian filter becomes the Kalman filter. 

The Bayesian filter consists of recursive prediction and the update steps. In the prediction step, it predicts a prior distribution of the current state ${\bf x_t}$ based on an old estimate using the state transition function, which is characterized by $p({\bf x_t}|{\bf z_{1:t-1}})$. In the update step, it updates prior state distribution to obtain a posterior estimate with new arrival measurements, which is characterized by $p({\bf x_t} | {\bf z_{1:t}})$.

\noindent{\bf Prediction Step}: 
The prior distribution $p({\bf x_t}|{\bf z_{1:t-1}})$ of the state is obtained using the Chapman-Kolmogorov equation and the state transition function,

\begin{equation}
p({\bf x_t}|{\bf z_{1:t-1}}) = \int p({\bf x_t} | {\bf x_{t-1}}) p({\bf x_{t-1}}|{\bf z_{1:t-1}})d{\bf x_{t-1}}. \label{eq:3}
\end{equation}

\noindent{\bf Updating Step}: When the new observation ${\bf z_t}$ is obtained, the prior distribution $p({\bf x_t}|{\bf z_{1:t-1}})$ is updated according to Bayes’ rule to estimate the posterior distribution $p({\bf x_t}|{\bf z_{1:t}})$,

\begin{equation}
     p({\bf x_t} | {\bf z_{1:t}}) = \frac{p({\bf z_t}|{\bf x_t}) p({\bf x_t}|{\bf z_{1:t-1}})}{p({\bf z_t} | {\bf z_{1:t-1}})}, \label{eq:4}
\end{equation}
where $p({\bf z_t}|{\bf z_{1:t-1}})$ can be expressed as,
\begin{equation}
p({\bf z_t}|{\bf z_{1:t-1}}) = \int p({\bf z_t}|{\bf x_t}) p({\bf x_t}|{\bf z_{1:t-1}}) d{\bf x_t}. \label{eq:5}
\end{equation}

RNNs bear resemblance to the Bayesian filters (see Fig.~{\ref{fig:3}}). Given a sequence of measurement, a Bayesian filter recursively estimates the optimal hidden states through two steps: prediction and update step, and optionally produce the target output every time step. Similarly, given a sequence of input, RNN updates the hidden state via a recurrent formula and optionally produces the output every time step.

The computation of RNNs is represented by the following equations,
\begin{align}
{\bf s_{t}} &= \varphi_s({\bf W_{ss}} {\bf s_{t-1}} +{\bf W_{zs}} {\bf z_{t-1}} + {\bf b_s}) \label{eq:6},\\
{\bf y_{t}} &= \varphi_o({\bf W_{sy}}{\bf s_t} + {\bf b_o}),
\end{align}
where ${\bf s_{t}}$ represents the hidden state at time step $t$, $\varphi_s$ and $\varphi_o$ are activation functions, ${\bf W_{ss}}$ is the hidden-to-hidden transformation matrix, ${\bf W_{zs}}$ is the input-to-hidden transformation matrix, ${\bf y_{t}}$ is the output, ${\bf W_{sy}}$ is the hidden-to-output transformation matrix, and ${\bf b_s}$ and ${\bf b_o}$ are the bias terms.

Gu \etal \cite{Gu_2017_CVPR} show Bayesian filters are a special type of RNNs with adaptive weights. While a Bayesian filter adapts its estimation models over time by changing weight, an RNN uses the fixed weight after training. Following this study, we assume an RNN performs the same procedure in updating the hidden state \eqref{eq:6} as do the two steps of the Bayesian filter.

The simple solution with which to deal with incomplete measurements in the Bayesian filter is an imputation. If we can impute missing data, we can use the Bayesian filtering steps in the same way. However, in the update step, the prior distribution is updated with the imputed value in \eqref{eq:4} and this might have a bad influence on the estimation of the hidden state. Furthermore, the model in which parameters are learned with imputed assumed data will affect the update step with actual measurements. Therefore, we explore the method that does not rely on imputation.

\begin{figure}[tb]
 \centering
 \includegraphics[width = \linewidth]{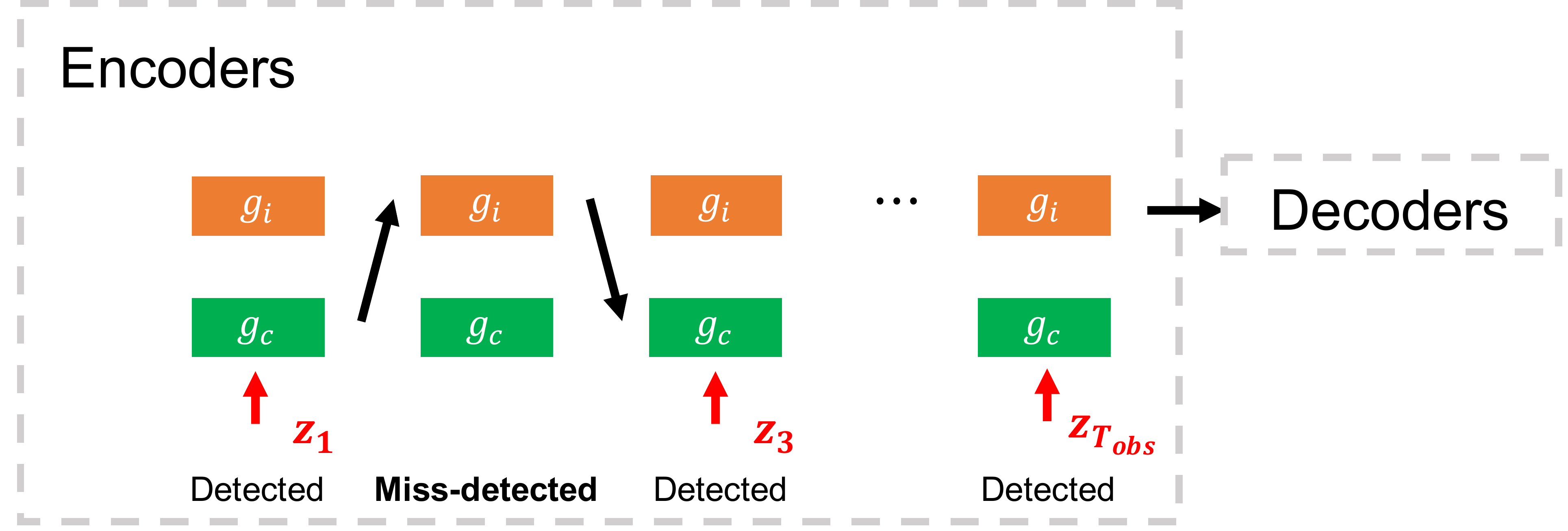}
 \caption{The flow of the proposed two-block RNN model. We use two functions, $g_c$ and $g_i$ for encoders depending on the object detection result. When miss-detection does not occur, we use $g_c$ as an encoder, and when miss-detection occurs, we use $g_i$ as an encoder.}
 \label{fig:4}
\end{figure}

\begin{figure}[tb]
 \centering
 \includegraphics[width =0.8 \linewidth]{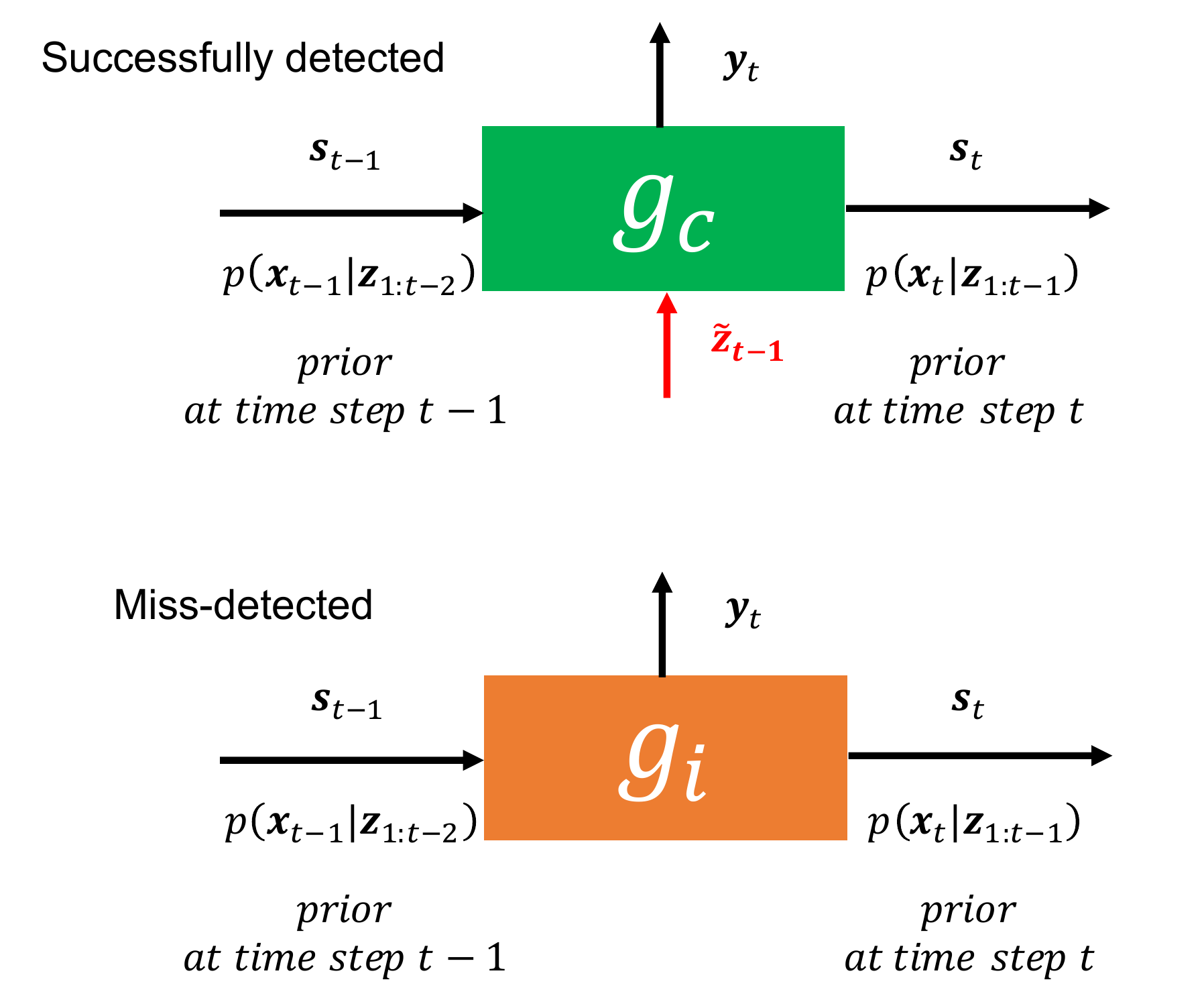}
 \caption{Two-block RNN diagram. The function $g_c$ approximates the update and prediction steps with the new measurement, and the function $g_i$ approximates the update and prediction steps without the new measurement.}
 \label{fig:5}
\end{figure}

\section{Proposed approach}

\subsection{Problem Definition}
We assume that the state of an agent is represented by its position and each scene is preprocessed by detection and tracking algorithms to obtain the position of each agent at each time instance. However, due to miss-detection, sometimes the obtained observed trajectory is incomplete. This includes the cases in which an agent is continuously detected in a sequence of frames but is not detected at the frame in the middle. The position at time step $t$ is denoted as ${\bf z_t}$, e.g., its 2D image coordinate in a bird's-eye view. When miss-detection does not occur, we can receive the complete trajectory ${\bf z_1},\ldots,{\bf z_{T_{obs}}}$. On the other hand, when miss-detection occurs, we can only access the incomplete observed trajectory. Our goal is to predict the probable future trajectory ${\bf z_{T_{obs}+1}},\ldots,{\bf z_{T_{pred}}}$ even from an incomplete observed trajectory. $T_{obs}$ and $T_{pred}$ denote the last observation and the last prediction time instance.
 
\subsection{Bayesian filter with miss-detection}

In the Bayesian filter, the future trajectory of the agent can be predicted with the appropriate hidden state and the observation functions. Thus, we can formulate the trajectory prediction task as the estimation of the hidden state ${\bf x_t}$, from which an output can be optionally derived, from all the measurements ${\bf z_{1:t-1}}=[{\bf z_1},\ldots,{\bf z_{t-1}}]$, which is represented by a probabilistic form: $p({\bf x_t}|{\bf z_{1:t-1}})$. However, in the case of miss-detection, we do not have access to all of ${\bf z}_t,t=1,\dots,T_{obs}$. To prevent the ambiguity in the notation, let us define $\tilde{{\bf z}}_{1:T_{obs}}$ to be the list of observed measurements until time step $T_{obs}$. Note that $\tilde{{\bf z}}_{1:T_{obs}}$ differs from ${\bf z}_{1:T_{obs}}$ since ${\bf z}_{1:T_{obs}}$ assumes all ${\bf z}_t$ up to time $T_{obs}$ to be observed, while for $\tilde{{\bf z}}_{1:T_{obs}}$ some ${\bf z}_t$ may be missing. With this notation, our goal becomes the estimation of the hidden state ${\bf x_{t}}$ from the incomplete data $\tilde{{\bf z}}_{1:t-1}$, which is represented in probabilistic form as $p({\bf x_{t}}|\tilde{{\bf z}}_{1:t-1})$, and can be used to predict the future trajectory.


In this case, one cycle of prediction and update step can be represented by the following derivation,
\begin{align}
p({\bf x_t}|\tilde{{\bf z}}_{1:t-1})
&= \int p({\bf x_t} | {\bf x_{t-1}}) p({\bf x_{t-1}}|\tilde{{\bf z}}_{1:t-1})d{\bf x_{t-1}}, \label{eq:BF_complete_int}\\
&= \int p({\bf x_t} | {\bf x_{t-1}}) p({\bf x_{t-1}}|\tilde{{\bf z}}_{1:t-2}, {\bf z}_{t-1})d{\bf x_{t-1}}, \label{eq:BF_complete_splitZ}\\
&= \frac{\int p({\bf x_t}|{\bf x_{t-1}}) p({\bf x_{t-1}}|\tilde{{\bf z}}_{1:t-2}) p({\bf z_{t-1}}|{\bf x_{t-1}})d{\bf x_{t-1}}}{p({\bf z_{t-1}}|\tilde{{\bf z}}_{1:t-2})}, \label{eq:BF_complete_bayesrule}\\
&= \frac{\int p({\bf x_t}|{\bf x_{t-1}}) p({\bf x_{t-1}}|\tilde{{\bf z}}_{1:t-2}) p({\bf z_{t-1}}|{\bf x_{t-1}})d{\bf x_{t-1}}}
{\int p({\bf z_{t-1}}|{\bf x_{t-1}}) p({\bf x_{t-1}}|\tilde{{\bf z}}_{1:t-2}) d{\bf x_{t-1}}}, \label{eq:11}
\end{align}
where in \eqref{eq:BF_complete_splitZ} we split the observed ${\bf z}_{t-1}$ from $\tilde{\bf z}_{1:t-1}$ in \eqref{eq:BF_complete_int}, then we have applied the Bayes' rule to obtain \eqref{eq:BF_complete_bayesrule}.

Here, the prior distribution $p({\bf x_{t-1}}|\tilde{{\bf z}}_{1:t-2})$ at time step $t-1$ is updated with the new observation $z_{t-1}$ to obtain the posterior distribution $p({\bf x_{t-1}}|\tilde{{\bf z}}_{1:t-1})$ at time step $t-1$. We can then estimate the prior distribution $p({\bf x_{t}}|\tilde{{\bf z}}_{1:t-1})$ at time step $t$ from this posterior distribution $p({\bf x_{t-1}}|\tilde{{\bf z}}_{1:t-1})$ at time step $t-1$. Notice that \eqref{eq:11} gives us a recurrent relation of computing $p({\bf x_t}|\tilde{{\bf z}}_{1:t-1})$ as a function of $p({\bf x_{t-1}}|\tilde{{\bf z}}_{1:t-2})$ and ${\bf z}_{t-1}$, which we will use as the foundation of our two-block model in Section \ref{sec:2block}.

To utilize the above mechanism when the miss-detection occurs, i.e., the new measurement ${\bf z_{t-1}}$ is not available, we conventionally need to synthetically generate a measurement by imputation for the update step \cite{Malla_2020_CVPR, Yao_2019_IROS}. However, updating the prior distribution with a synthetically generated measurement might have a bad influence on the estimation of hidden state ${\bf x_{t}}$. To avoid the wrong update, we estimate the hidden state ${\bf x_{t}}$ from the measurements $\tilde{{\bf z}}_{1:t-2}$ instead of compensating for the missing data with synthetically generated data,
\begin{align}
p({\bf x_t}|\tilde{{\bf z}}_{1:t-1}) 
&= p({\bf x_t}|(\tilde{{\bf z}}_{1:t-2}, {\bf z}_{t-1} \text{ not observed})),\\
&= p({\bf x_t}|\tilde{{\bf z}}_{1:t-2}),\\
&= \int p({\bf x_t} |{\bf x_{t-1}}) p({\bf x_{t-1}}|\tilde{{\bf z}}_{1:t-2})d{\bf x_{t-1}}.\label{eq:BF_incomplete_rec}
\end{align}
We can see that the prior distribution $p({\bf x_{t}}|\tilde{{\bf z}}_{1:t-1})$ at time step $t$ can be directly estimated from the prior distribution $p({\bf x_{t-1}}|\tilde{{\bf z}}_{1:t-2})$ at time step $t-1$ without the update step. This elimination of the update step prevents the model from accumulating the error that is caused by a wrong update every time step. Notice again that \eqref{eq:BF_incomplete_rec} provides a recurrent relation to compute $p({\bf x_{t}}|\tilde{{\bf z}}_{1:t-1})$ as a function of $p({\bf x_{t-1}}|\tilde{{\bf z}}_{1:t-2})$, but without the observation ${\bf z}_{t-1}$. Using these recurrent relations, we can develop an algorithm for handling incomplete trajectories as described in the next section.
 
 \begin{table*}
 \caption{Quantitative results of all methods across different datasets. We present the results on two evaluation metrics (ADE and FDE)} for the task of predicting $8$ and $12$ future time steps, given the $8$ previous ones. Best results are highlighted in bold.
\label{table:2}
\begin{center}

\begin{tabular}{ccc@{\hspace*{3mm}}c@{\hspace*{3mm}}c@{\hspace*{3mm}}c@{\hspace*{3mm}}c@{\hspace*{6mm}}c@{\hspace*{3mm}}c@{\hspace*{3mm}}c@{\hspace*{3mm}}c@{\hspace*{3mm}}c}
\toprule
\multirow{3.5}{*}{Metric} & \multirow{3.5}{*}{Dataset} & \multicolumn{5}{c}{Trained to predict 8 future time steps} & \multicolumn{5}{c}{Trained to predict 12 future time steps} \\ \cmidrule(l{2mm}r{5mm}){3-7} \cmidrule(l{-1.5mm}){8-12} 
 &  & Last Fill. & Zero Fill. & Linear Fill. & \begin{tabular}[c]{@{}c@{}}Two-Block\\ RNN (Ours)\end{tabular} & \begin{tabular}[c]{@{}c@{}}Complete\\ Data \end{tabular} & Last Fill. & Zero Fill. & Linear Fill. & \begin{tabular}[c]{@{}c@{}}Two-Block\\ RNN (Ours)\end{tabular} & \begin{tabular}[c]{@{}c@{}}Complete\\ Data \end{tabular} \\ 
 \midrule
\multirow{6.6}{*}{ADE} & ETH & 1.57 & 1.26 & 1.16 & \textbf{0.97} &0.91  &1.87  & 1.78  &1.50  &\textbf{1.41}  & 1.35 \\
 & HOTEL & 4.41 & 4.70 & 3.57 & \textbf{3.49} & 3.10 &5.12  & 4.84 & 4.60 & \textbf{4.56} & 6.63\\
 & UNIV & 1.17 & 1.33 & 1.11 & \textbf{1.02} & 0.95 & 1.61  & 2.04 & 1.66 & \textbf{1.53} & 1.40 \\
 & ZARA1 & 0.62 & 0.65 & 0.66 & \textbf{0.52} & 0.40 & 0.89 & 0.95 & 0.89 & \textbf{0.78} & 0.68\\
 & ZARA2 & 0.68 & 0.71 & 0.78 & \textbf{0.52} & 0.40 & 1.00 & 1.07  & 1.10 & \textbf{0.83} & 0.63 \\
 \cmidrule(l){2-12} 
  & AVG & 1.69 &1.73  & 1.46 &  \textbf{1.30} & 1.15 & 2.10 & 1.94 &1.95 & \textbf{1.82} & 2.14 \\
 \midrule
\multirow{6.5}{*}{FDE} & ETH &  2.78 & 2.30 & 2.10 & \textbf{1.81} & 1.74 & 3.14 & 3.23 &  2.69& \textbf{2.61} & 2.57 \\
 & HOTEL & 5.17 & 5.86 &4.66  & \textbf{4.39} & 3.89 & 5.94 & 5.61 &\textbf{5.00} & 5.50 &8.46 \\
 & UNIV & 2.06 & 2.18 &1.92  & \textbf{1.82} & 1.67 & 2.92 & 3.35 & 2.97 & \textbf{2.80} & 2.56\\
 & ZARA1 & 1.09 & 1.12 & 1.16 & \textbf{0.93} & 0.74 & 1.59 & 1.70  & 1.57 & \textbf{1.46} & 1.30 \\
 & ZARA2 & 1.21 & 1.25 &1.43  & \textbf{0.95} &0.74  &1.98  & 2.03 & 2.14  & \textbf{1.56} & 1.23 \\
 \cmidrule(l){2-12} 
  & AVG & 2.46  &2.54  & 2.25 & \textbf{1.98} &1.76  & 3.11 & 3.18  & 2.87 &  \textbf{2.79} & 3.22 \\
 \bottomrule
\end{tabular}

\end{center}

\end{table*}

\subsection{Two-block model for encoding miss-detection}
\label{sec:2block}

Inspired by the connection between RNNs and the Bayesian filter \cite{Gu_2017_CVPR}, we apply the two above recurrent relations in the Bayesian filter with RNNs. Similar to the cases in the Bayesian filter, we use two RNNs depending on the detection result (see Fig. {\ref{fig:4})}. One is used when the new measurement is available, and another is used when miss-detection occurs and the new measurement is not available for avoiding the wrong update. Fig.{\ref{fig:5} shows a diagram of our two-block RNN model.}
 
When the agent is successfully detected and a new measurement is available, we assume that one RNN works as a function that approximates a function approximating \eqref{eq:11}, which estimates the prior distribution at time step $t$ from the prior distribution at time step $t-1$ and the new measurement,
\begin{align}
{\bf s_{t}}
&= g_c({\bf s_{t-1}},{\bf z_{t-1}}),\label{eq:RNNcomplete}
\end{align}
where $g_c$, implemented as an RNN, is a function that approximates the update and prediction steps with a new measurement in \eqref{eq:11}. Note the similarity between \eqref{eq:11} and \eqref{eq:RNNcomplete}: the former is a recurrent relation between $p({\bf x_t}|\tilde{{\bf z}}_{1:t-1})$, $p({\bf x_{t-1}}|\tilde{{\bf z}}_{1:t-2})$, and ${\bf z}_{t-1}$, while the latter is a recurrent relation between ${\bf s}_t$, ${\bf s}_{t-1}$, and ${\bf z}_{t-1}$. Thus, we can interpret \eqref{eq:RNNcomplete} as representing $p({\bf x_t}|\tilde{{\bf z}}_{1:t-1})$ by ${\bf s}_t$ and $g_c$ being a function that approximates the computation of the expression in \eqref{eq:11}.

When miss-detection occurs and the new measurement is not available, we assume another RNN works as a function approximating \eqref{eq:BF_incomplete_rec}, which directly estimates the prior distribution at time step $t$ from the prior distribution at time step $t-1$ without a new measurement:
\begin{align}
{\bf s_{t}}
&= g_i({\bf s_{t-1}}),\label{eq:RNNincomplete}
\end{align}
where $g_i$, implemented as an RNN, is a function that approximates the prediction step without the new measurement and the update step, as in \eqref{eq:BF_incomplete_rec}. Again, one could see a similar relationship between \eqref{eq:BF_incomplete_rec} and \eqref{eq:RNNincomplete}, which allows us to draw a similar interpretation to the case between \eqref{eq:11} and \eqref{eq:RNNcomplete}.

\subsection{Prediction}
In the previous section, we encode the incomplete observed trajectory using the two-block RNN depending on the detection result. In order to predict the future trajectory $\hat{{\bf z}}_{T_{obs+1}:T_{pred}}$ using the encoded information, we use another RNN:
\begin{align}
{\bf s_{t}}
&= g_p({\bf s_{t-1}}, \hat{{\bf z}}_{t-1}), \\
 \hat{{\bf z}}_{t}
&= h_p({\bf s_{t}}),
\end{align}
where $g_p$ is a function that approximates a cycle of update and prediction steps, and $h_p$ is a function that approximates an observation function that predicts the measurement from the hidden state (implemented as a Multilayer Perceptron (MLP) in our experiments). The pseudocode for the algorithm is provided in Alg. \ref{alg:2blockrnn}.

\begin{algorithm}
 \caption{Two-block RNN model for future trajectory prediction \label{alg:2blockrnn}}
 \begin{algorithmic}[1]
 \renewcommand{\algorithmicrequire}{\textbf{Input:}}
 \renewcommand{\algorithmicensure}{\textbf{Output:}}
 \REQUIRE Incomplete observed states of the agent $\tilde{{\bf z}}_{1:T_{obs}}$
 \ENSURE   Complete future states of the agent $\hat{{\bf z}}_{T_{obs+1}:T_{pred}}$
  \FOR {$t = 1$ to $T_{obs}$}
  \IF {the target agent is miss-detected}
  \STATE ${\bf s_{t}} = g_i({\bf s_{t-1}})$;
  \ELSE
  \STATE ${\bf s_{t}} = g_c({\bf s_{t-1}},{\bf z_{t-1}})$;
  \ENDIF
  \ENDFOR
  \FOR {$t = T_{obs+1}$ to $T_{pred}$}
  \STATE ${\bf s_{t}}= g_p({\bf s_{t-1}}, \hat{{\bf z}}_{t-1})$;
  \STATE  $\hat{{\bf z}}_{t} = h_p({\bf s_{t}})$;
  \ENDFOR
 \RETURN $\hat{{\bf z}}_{T_{obs+1}:T_{pred}}$
 \end{algorithmic} 
 \end{algorithm}

  \begin{figure*}[tb]
\begin{center}
{\begin{tabular}{cc}
\begin{minipage}{0.45\hsize}
    \begin{center}
        \includegraphics[trim={0 1em 0 0}, clip, width=\hsize]{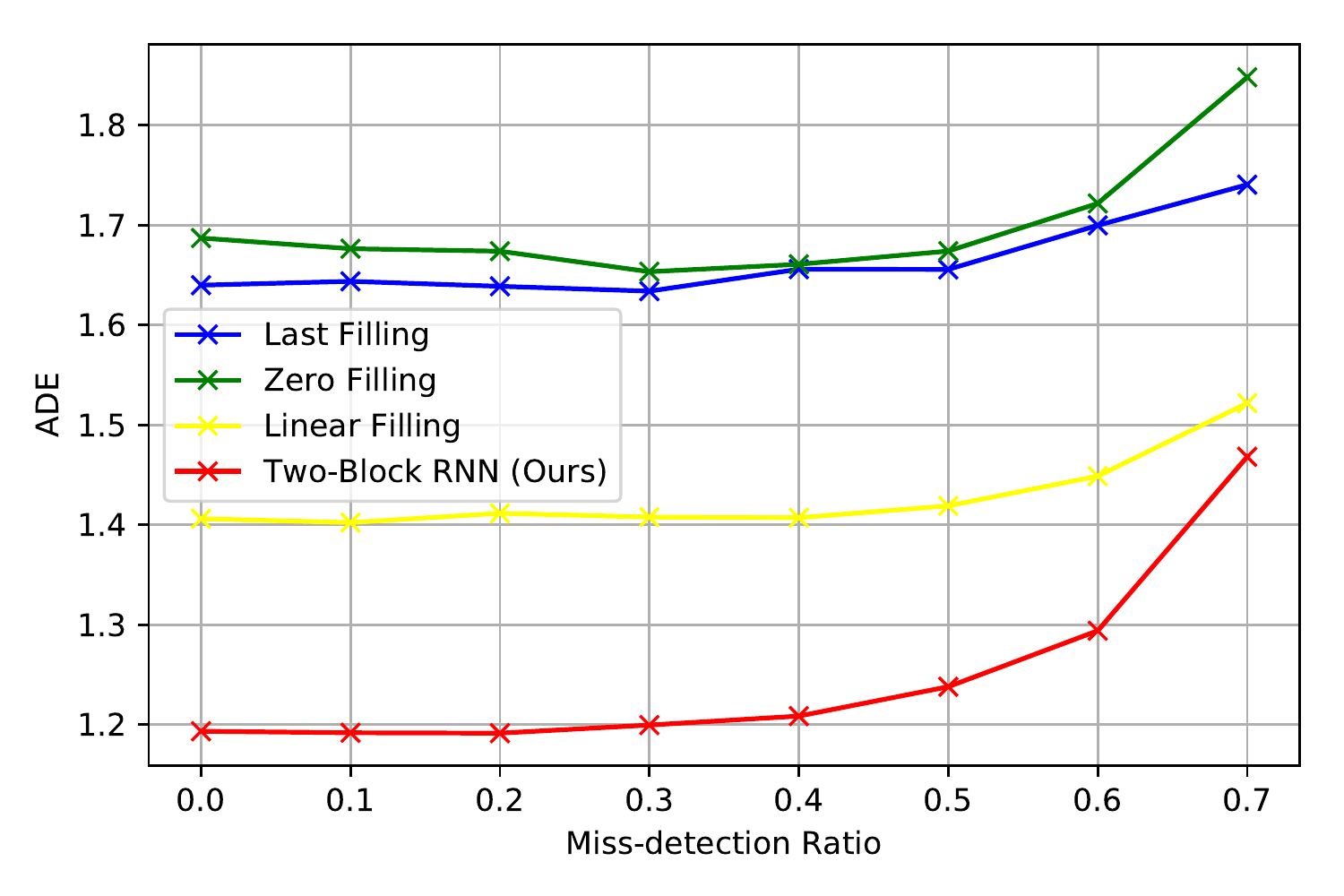} 
        {\footnotesize (a) ADE of models trained to predict 8 future time steps}
    \end{center}
\end{minipage}
&
\begin{minipage}{0.45\hsize}
    \begin{center}
        \includegraphics[trim={0 1em 0 0}, clip, width=\hsize]{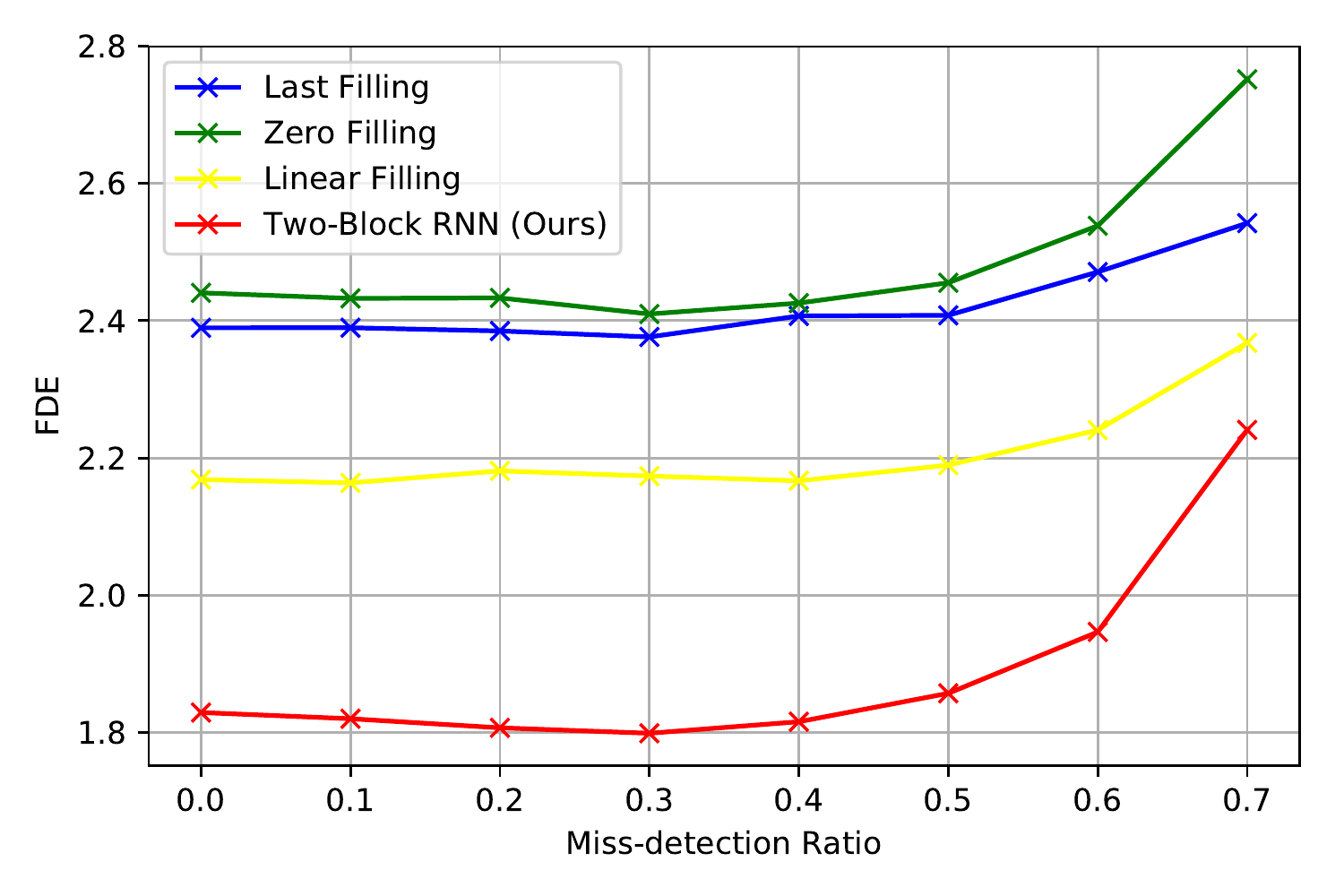}
        {\footnotesize (b) FDE of models trained to predict 8 future time steps}
    \end{center}
\end{minipage}
\vspace{0.5em}
\\
\begin{minipage}{0.45\hsize}
    \begin{center}
        \includegraphics[trim={0 1em 0 0}, clip, width=\hsize]{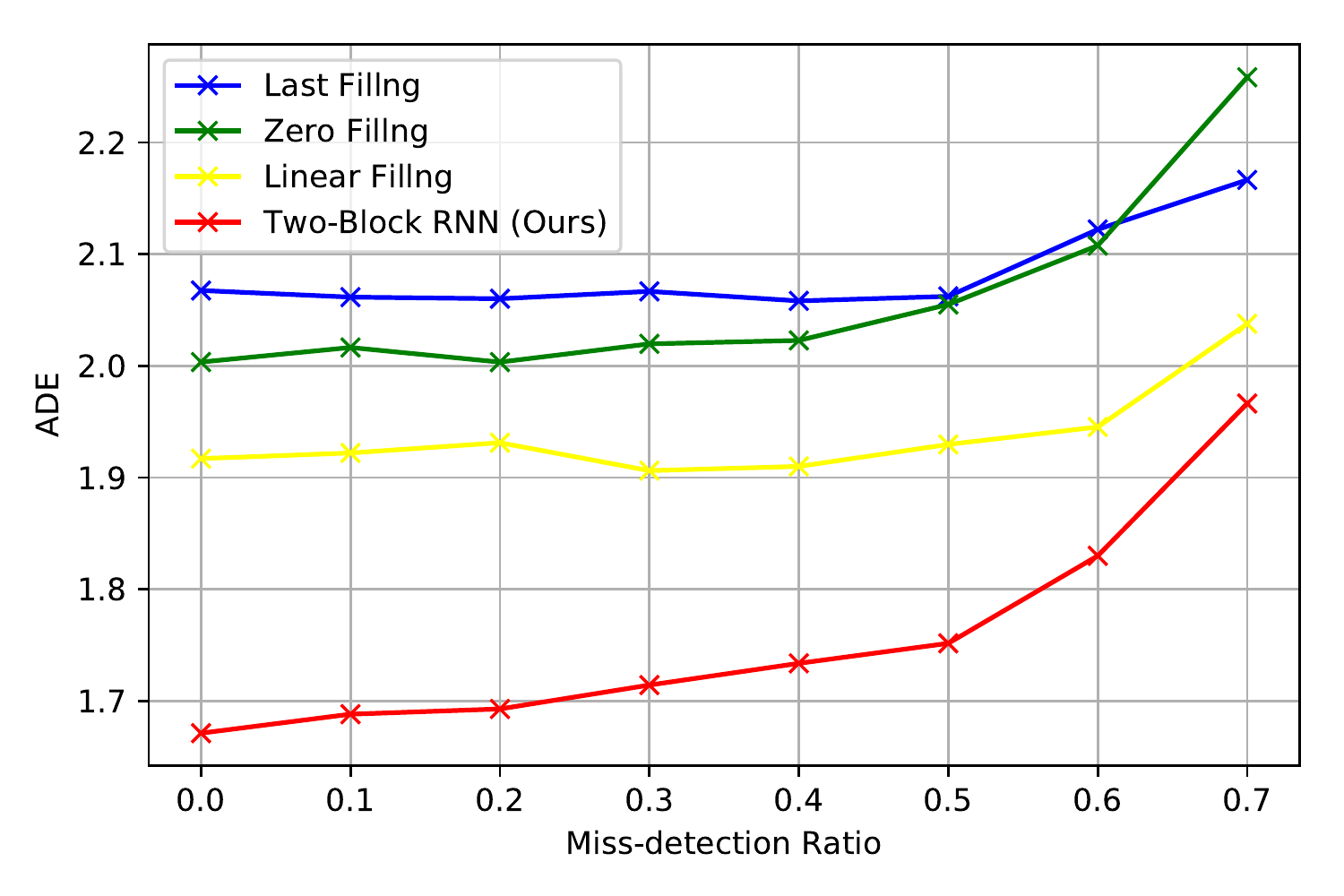} 
        {\footnotesize (c) ADE of models trained to predict 12 future time steps}
    \end{center}
\end{minipage}
&
\begin{minipage}{0.45\hsize}
    \begin{center}
        \includegraphics[trim={0 1em 0 0}, clip, width=\hsize]{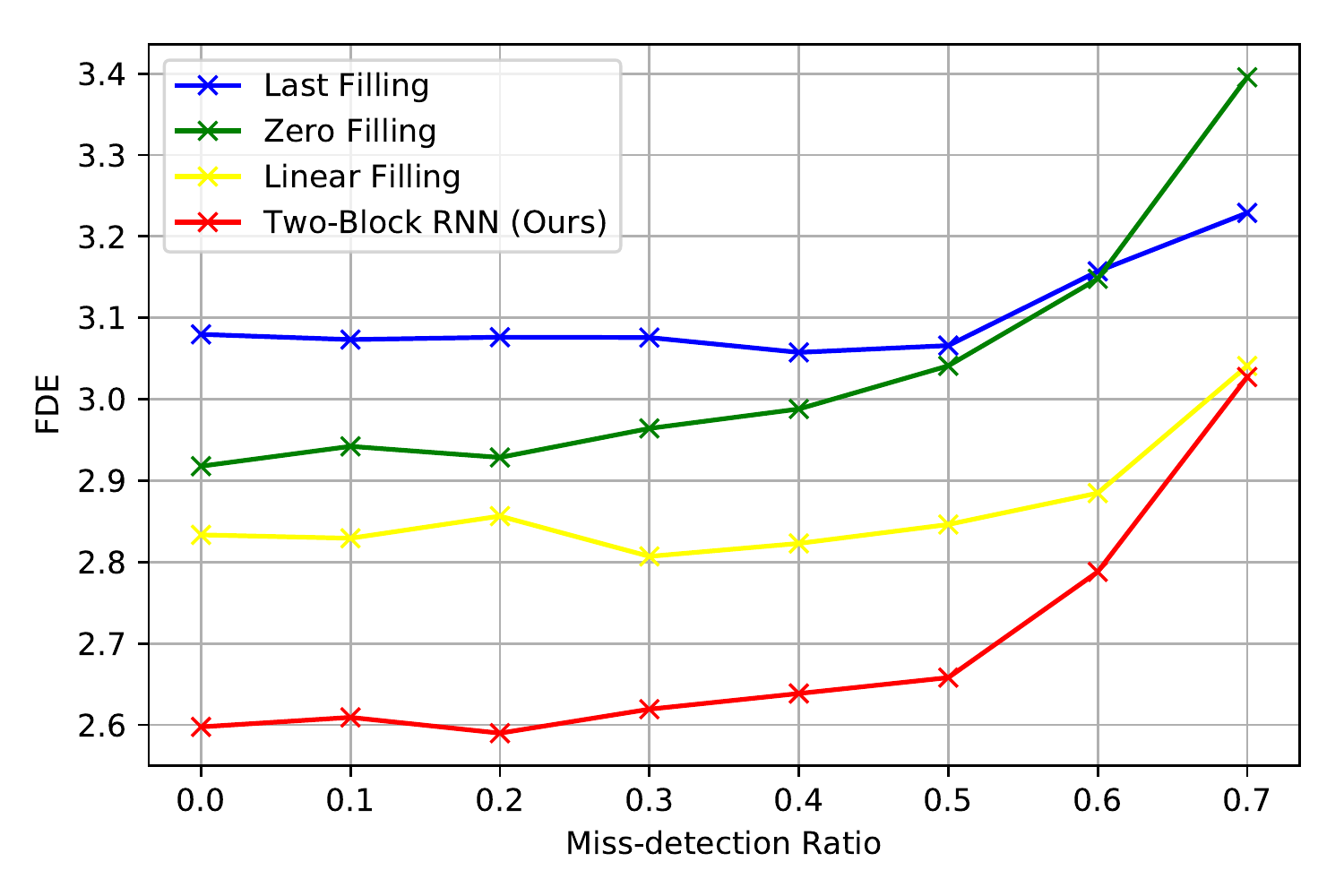}
        {\footnotesize (d) FDE of models trained to predict 12 future time steps}
    \end{center}
\end{minipage}
\vspace{0.5em}
\end{tabular}}
\caption{We compare the robustness of our proposed model and baselines against the miss-detection ratio changes on ADE and FDE metrics. Our two-block RNN model outperforms the baselines in every miss-detection ratio. Our method also achieves superior performance when no miss-detection occurs. 
The results are averaged over all datasets.}
\label{fig:6}
\end{center}
\end{figure*}

\section{Experiments}
In this section, we perform experiments to evaluate our two-block RNN model against several imputation baseline approaches. All experiments are performed on a computer with an Intel Core Xeon CPU and an Nvidia Tesla K80 GPU. We use PyTorch~\cite{NEURIPS2019_9015} for our implementation.

\subsection{Experimental settings}
To evaluate our two-block RNN method, we experiment with the existing trajectory prediction model, Social GAN \cite{Gupta_2018_CVPR}, a method to retrieve multiple possible future paths for multi-agents, on two publicly available datasets: ETH~\cite{Pellegrini_2010_ECCV} and UCY~\cite{Leal_2014_CVPR}. We follow the same setting of Social GAN. For our appoach, we only modify the encoders of Social GAN, which use a single agent encoder to encode the observed measurements of each agent independently into our two-block RNN.

{\bf Baselines.} We compare the performance of the proposed method with three baselines. The incomplete data due to miss-detection cannot be directly inputed into Social GAN. We firstly impute incomplete data for generating the synthetic complete data and then input the data to the model. We compare against the following imputation methods:
 \begin{itemize}
  \item {\bf Last filling}: Impute the missing measurement with the last detected measurement.
  \item {\bf Zero filling}: Impute the missing measurement with zero.
  \item {\bf Linear filling}: Interpolate the missing measurement linearly using previous and next observed measurements. When the miss-detection happens at the end of the observation time, we extrapolate the missing measurement linearly.
 \end{itemize}

{\bf Implementation details.} Social GAN uses LSTMs as the RNNs in their model for encoders and sets $32$ as the dimension of the hidden states for the encoder. We also use the LSTMs in our two-block RNN model. We halved the size of the hidden state of our two-block model, which consists of two RNNs for fair comparison in terms of the number of parameters. For other settings, we followed the original Social GAN.

As for the representation of measurement, Social GAN utilizes relative coordinates for translation  invariance. However, when miss-detection occurs and the position at a time step is not available, we cannot compute both the previous and next relative coordinates around the time step. Therefore, we use absolute coordinates instead of relative coordinates. We normalize the pedestrian positions by subtracting the mean and dividing by the standard deviation of the training set.

\begin{figure*}[tb]
\begin{center}
\begin{tabular}{cc}
\begin{minipage}{0.45\hsize}
    \begin{center}
        \includegraphics[trim={0 1em 0 0}, clip, width=\hsize]{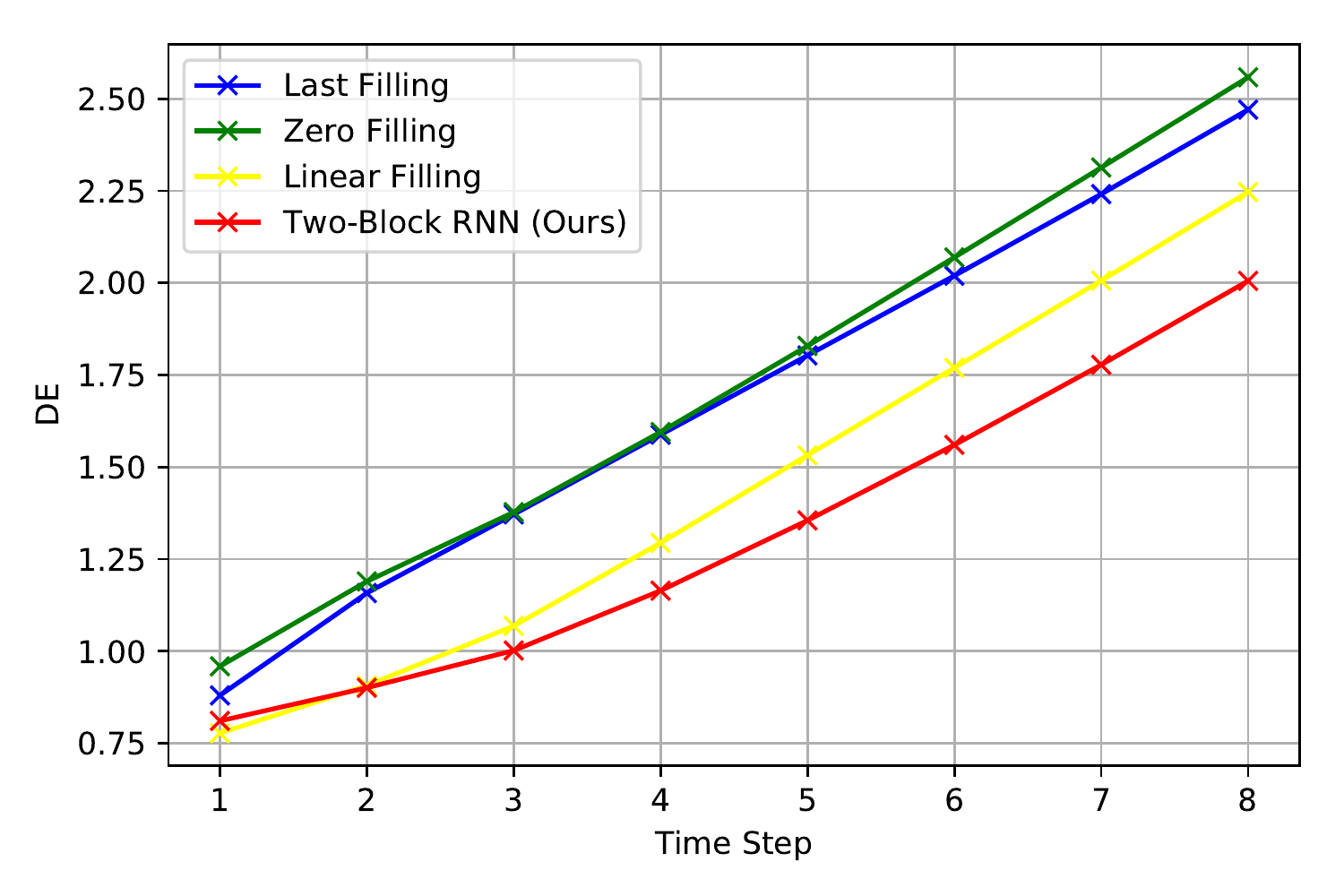} 
        {\footnotesize (a) DE of models trained to predict 8 future time steps}
    \end{center}
\end{minipage}
&
\begin{minipage}{0.45\hsize}
    \begin{center}
        \includegraphics[trim={0 1em 0 0}, clip, width=\hsize]{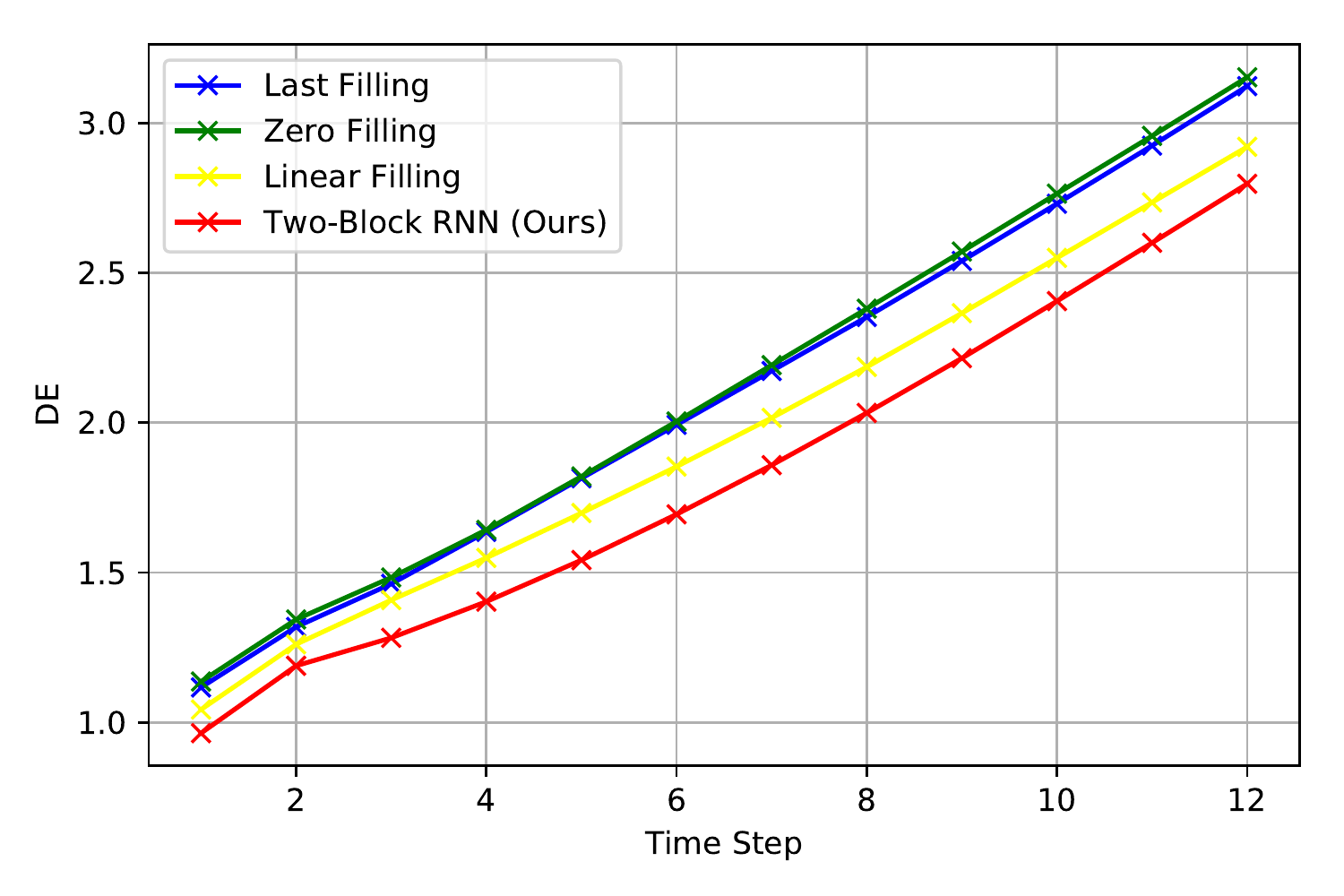}
        {\footnotesize (b) DE of models trained to predict 12 future time steps}
    \end{center}
\end{minipage}
\end{tabular}
\caption{Displacement error at each time step. Our model provides better prediction quality over all time steps. The results are averaged over all datasets.}
\label{fig:7}
\end{center}
\end{figure*}

{\bf Datasets.} 
We evaluate our two-block RNN model on two datasets: ETH \cite{Pellegrini_2010_ECCV} and UCY \cite{Leal_2014_CVPR}, which are commonly used for the trajectory prediction task. The ETH dataset contains two scenes named ETH and HOTEL. UCY dataset includes three scenes named ZARA-01, ZARA-02, and UCY. They are collected from bird’s-eye views containing thousands of real-world pedestrian trajectories, and covering numerous challenging situations. We observe trajectories for 3.2 seconds (8 frames) and predict for 3.2 seconds (8 frames) or 4.8 seconds (12 frames), and use a leave-one-out approach, i.e., train on four sets and test on the remaining set for evaluation. 

Unfortunately, ETH and UCY do not provide a miss-detection label. Therefore, we synthetically generate miss-detection masks. To generate the miss-detection masks, we randomly choose a miss-detection ratio from $[0.2, 0.8]$ for every sequence both in training and evaluation. 

{\bf Evaluation metrics.} We follow the prior works  \cite{Alahi_2016_CVPR, Gupta_2018_CVPR} for evaluation metrics. We use the Average Displacement Error (ADE) and the Final Displacement Error (FDE) metrics. ADE is the average $L2$ distance between the predictions and the ground truth overall predicted time step, and the FDE is the $L2$ distance between the prediction and ground truth of final destination.

\begin{table*}[tb]
\caption{Results of the evaluation on different prediction lengths from models trained to predict 8 and 12 future time steps. Best results are highlighted in bold. The results are averaged over all datasets.}
\label{table:3}
\begin{center}
\begin{tabular}{cccccccccc}
\toprule
\multirow{3.5}{*}{Metric} & \multirow{3.5}{*}{\begin{tabular}[c]{@{}c@{}}Pred. \\ Length\end{tabular}} & \multicolumn{4}{c}{Trained to predict 8 future time steps} & \multicolumn{4}{c}{Trained to predict 12 future time steps} \\ \cmidrule(l{2pt}r{3pt}){3-6} \cmidrule(l{2pt}){7-10} 
 &  & Last Fill. & Zero Fill. & Linear Fill. & \begin{tabular}[c]{@{}c@{}}Two-Block\\ RNN (Ours)\end{tabular} & Last Fill. & Zero Fill. & Linear Fill. & \begin{tabular}[c]{@{}c@{}}Two-Block\\ RNN (Ours)\end{tabular} \\ \midrule
\multirow{3}{*}{ADE} & 12 & 3.03 & 3.20 & 2.73 & \textbf{2.47} &2.09   & 2.13 & 1.97 & \textbf{1.83}  \\
 & 16 & 4.85 & 5.04 & 4.47 & \textbf{3.97} & 3.33  & 3.19 & 3.02 & \textbf{2.81}  \\
 & 20 & 7.48 & 7.54 & 6.98 & \textbf{6.13} & 4.96 & 4.72 & 4.41 & \textbf{4.17}   \\
 \midrule
\multirow{3}{*}{FDE} & 12 & 3.10 & 3.32  & 2.97 & \textbf{2.69} & 3.10 & 3.16 & 2.92 & \textbf{2.80}  \\
 & 16 & 3.95 & 4.13 & 3.84 & \textbf{3.37} & 3.82  & 3.70 & 3.50 & \textbf{3.32}  \\
 & 20 & 5.16  & 5.27  & 5.16 & \textbf{4.35} & 4.80 & 4.56 & 4.30 & \textbf{4.13} \\ \bottomrule
\end{tabular}

\end{center}

\end{table*}

\subsection{Experimental Results}
{\bf Main Results.} In Table \ref{table:2}, we evaluate our model against all baseline models. We see that the linear imputation baseline that linearly estimates the missing state outperforms the other baselines, which do not estimate the missing state. Our two-block RNN model outperforms baselines on almost all of the datasets on both metrics. The best result among the baselines on the ADE metric is linear imputation with an error of $1.46$ ($T_{pred}=8$) and $1.95$ ($T_{pred}=12$). Our model has the ADE error of $1.30$ ($T_{pred}=8$) and $1.82$ ($T_{pred}=12$), which is $11\%$ and $6\%$ less than the best baseline, respectively, and the effect becomes the most prominent on ZARA2 ($33\%$ in $T_{pred}=8$ and $25\%$ in $T_{pred}=12$). In terms of the FDE metric, we can also get the improvement of $12\%$ ($T_{pred}=8$) and $2\%$ ($T_{pred}=12$), compared to the best baseline. 

{\bf Changing miss-detection ratio.} We compare the robustness of the two-block RNN model and baselines when the miss-detection ratio changes. We test the model, which is trained with a random miss-detection ratio chosen from $[0.2, 0.8]$ under various fixed miss-detection ratios. In Fig. \ref{fig:6}, we vary the miss-detection ratio, from $0$ (complete data) to $0.7$. Overall, our two-block model can get better results compared to baselines in every miss-detection ratio setting. The performance becomes better as the miss-detection ratio becomes lower. When no miss-detection occurs, i.e., the miss-detection ratio is $0$, our model achieves the best performance compared to the baseline model on both ADE and FDE metrics, in both $8$ and $12$ prediction time-step settings. In the baselines, a single RNN is trained to encode both real data and imputed data. As imputed data may be confused with measurement, this could cause the model to learn the incorrect values, leading to high-error prediction even when there is no miss-detection in test time. In the $8$-time-step prediction, our model outperforms the best baseline model, linear filling by $15\%$ (ADE) and $15\%$ (FDE). In $12$-time-step prediction, our model achieves superior performance against the best baseline model, linear filling by $12\%$ (ADE) and $8\%$ (FDE). On the other hand, in our two-block RNN model, one RNN, namely, $g_c$ is trained to encode the real data and another RNN, namely, $g_i$ is trained to update the hidden state without a new measurement. Therefore, when no miss-detection occurs in test time, we can use $g_c$ which is trained only with real data as encoders, and this leads to better performance compared to the baselines. 

\begin{table}[tb]
\caption{Computation time comparison (in milliseconds) for 8-time-step prediction. The means and the standard deviations of the computational time are reported.  
}
\label{table:5}
\begin{center}
\begin{tabular}{c@{\hspace*{3mm}}c@{\hspace*{3mm}}c@{\hspace*{3mm}}c@{\hspace*{3mm}}c}
\toprule
\multirow{2.5}{*}{Method} & \multicolumn{4}{c}{Computation time (milliseconds)} 
\\
\cmidrule(l{-1mm}){2-5}
 & Filling & Encoding &  Prediction & Total
 \\
  \midrule
  Last Fill. &0.22 $\pm$ 0.07 & 0.49 $\pm$ 0.10  & 3.94 $\pm$ 0.55 & 4.66 $\pm$ 0.72
  \\
Zero Fill. & 0.03 $\pm$ 0.01   & 0.49 $\pm$ 0.09 & 4.01 $\pm$ 0.53 & 4.55 $\pm$ 0.60
 \\
Linear Fill. & 16.0 $\pm$ 17.5  & 0.47 $\pm$ 0.08  & 3.96 $\pm$ 0.37 & 20.3 $\pm$ 17.7
 \\
Ours & - & 8.02 $\pm$ 0.86 & 3.72 $\pm$ 0.24 &  11.74 $\pm$ 0.98
\\
\bottomrule
\end{tabular}
\end{center}

\end{table}

{\bf Displacement error (DE) at each time step.} 
In Fig. \ref{fig:7}, we report the displacement error at each prediction time step of all methods. Our model performs better every time step in both $8$ and $12$ prediction time step settings. This result suggests better prediction quality over all the time instants. 

{\bf Changing prediction length.} 
To show the stability in predicting longer temporal horizons, we present the ADE and FDE for the prediction of $12$, $16$, and $20$ future time steps in Table \ref{table:4}. Here, we use the same models trained to predict $8$ and $12$ time steps as in the main results, and simply extend their prediction time steps. The performance becomes worse as the prediction length becomes longer. Still, our model has a consistent advantage at every prediction time-step setting.

{\bf Aligning the hidden state channels.}
To provide a fair comparison of our model, in the main results, we halve the number of channels of the hidden state of our two-block model, which consists of two RNNs (each having $16$ channels), so that the total number of channels of our two-block model is the same as that of the baselines ($32$ channels). In this section, we run an additional experiment in which we align and vary the number of channels of both our model and the baseline RNNs, so that all RNNs, both of our model and the baselines, have the same number of channels, in order to confirm that the performance of our model is not caused by the difference in the number of channels. 
The results are shown in Table \ref{table:4}. 
We can see that our model can still outperform the baselines.
Hence, we show that the performance of our model is not caused by the difference in the number of hidden state channels.

{\bf Computation time.}
Table {\ref{table:5}} shows the comparison of the computation time of our method and the baseline methods. The total time is broken down into the time for filling missing values, encoding the trajectories, and making predictions. 
Note that the baselines require the filling time to fill the missing values, while our method does not. 
In terms of total time, last filling and zero filling are the fastest, followed by our method and then by linear filling.  
Looking at the breakdown, we can see all methods require roughly the same time for prediction, since they all use the same implementation. 
On the other hand, linear filling requires more time because it takes longer to compute the imputed values. 
Our method requires more time than the baselines do for encoding the trajectories.
This is because our method requires selecting one LSTM from the two LSTM blocks to pass the data to in each time step\footnote{For our method's implementation in this work, we simply put the two-block model in a \texttt{for} loop and iterate over the time steps.}, and thus it does not receive the speed-up benefit from the optimized LSTM implementation used by the baselines.
However, since the selection only requires an additional \texttt{if} statement, an optimized implementation of our two-block method should be able to achieve almost the same computation time as those of the baselines.

\begin{figure*}[tb]
\begin{center}
\begin{tabular}{ccccc}
{\small (a)}
&
\begin{minipage}{0.22\hsize}
    \begin{center}
    {\footnotesize\bf Last Filling}
        \includegraphics[clip, width=\hsize]{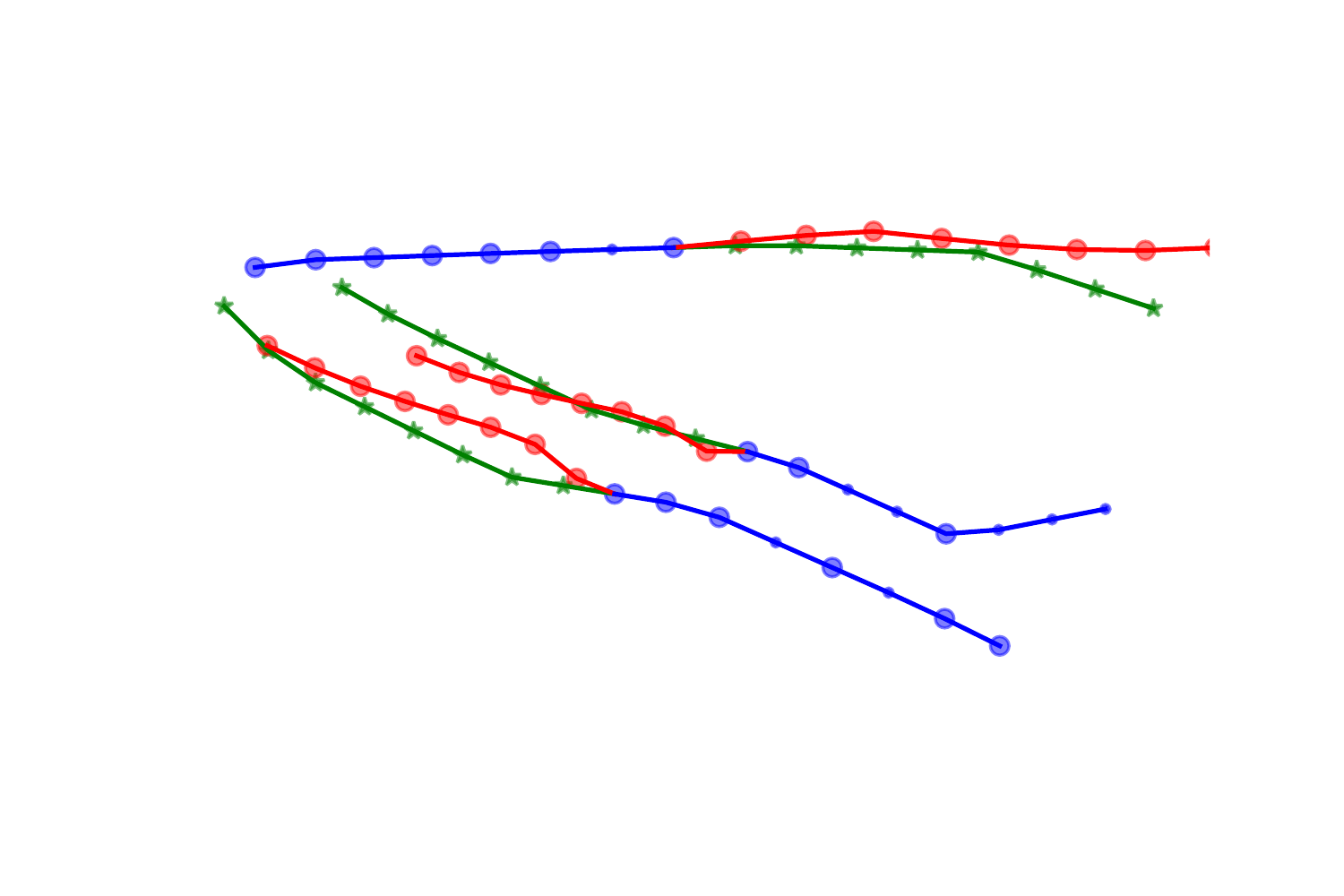} 
    \end{center}
\end{minipage}
&
\begin{minipage}{0.22\hsize}
    \begin{center}
    {\footnotesize\bf Zero Filling}
        \includegraphics[clip, width=\hsize]{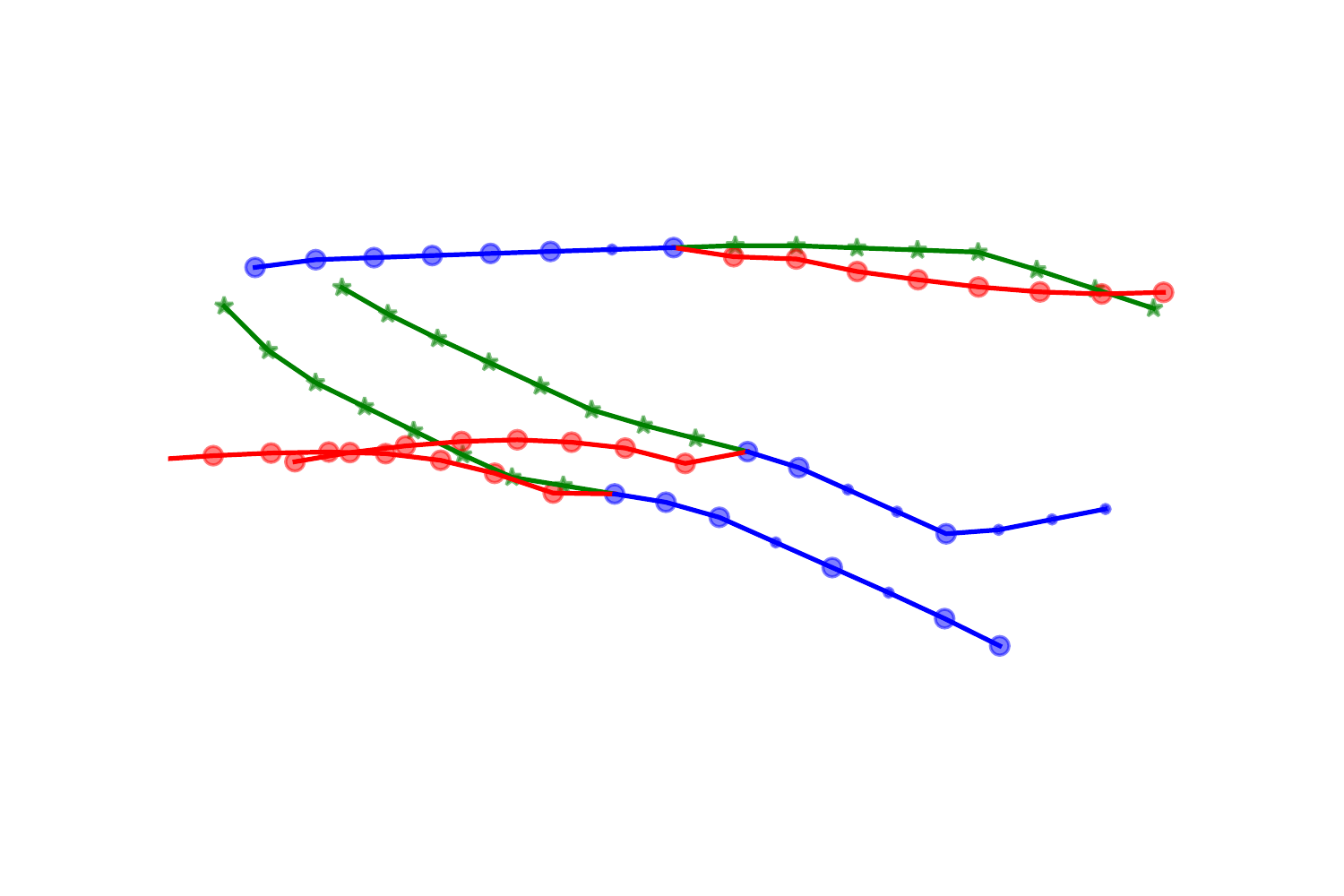}
    \end{center}
\end{minipage}
&
\begin{minipage}{0.22\hsize}
    \begin{center}
    {\footnotesize\bf Linear Filling}
        \includegraphics[clip, width=\hsize]{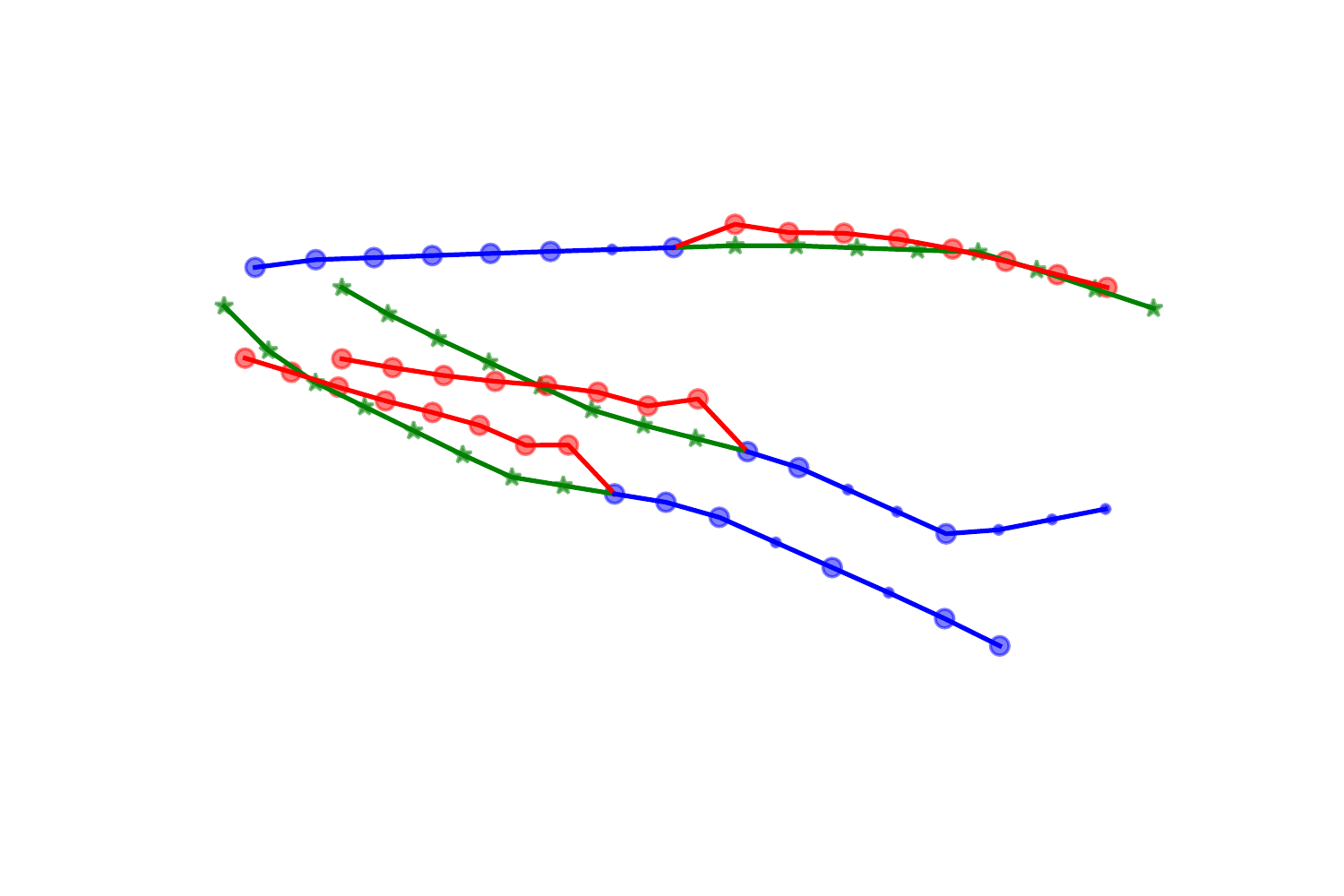}
    \end{center}
\end{minipage}
&
\begin{minipage}{0.22\hsize}
    \begin{center}
    {\footnotesize\bf Two-Block RNN (Ours)}
        \includegraphics[clip, width=\hsize]{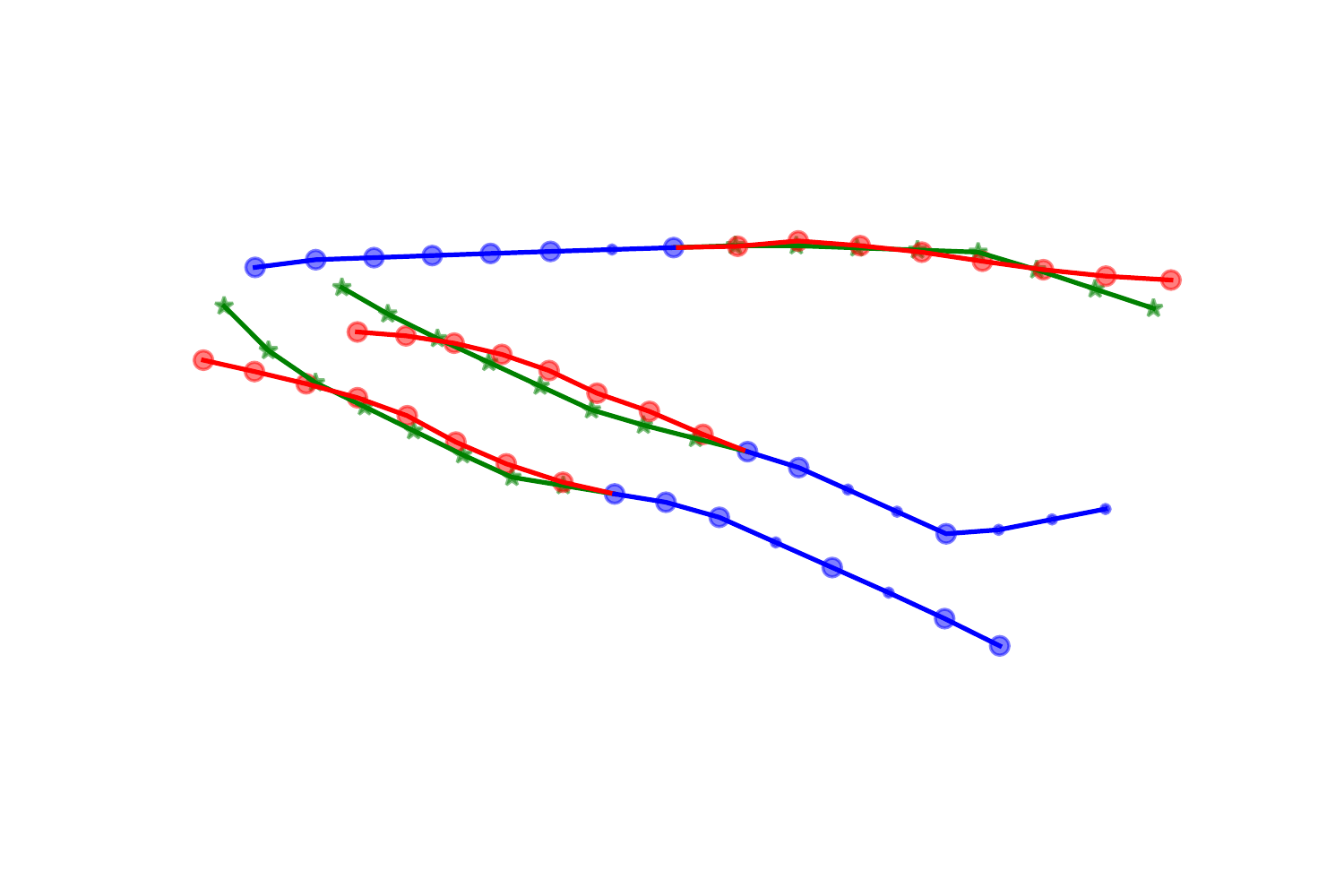}
    \end{center}
\end{minipage}
\\
{\small (b)}
&
\begin{minipage}{0.22\hsize}
    \begin{center}
        \includegraphics[clip, width=\hsize]{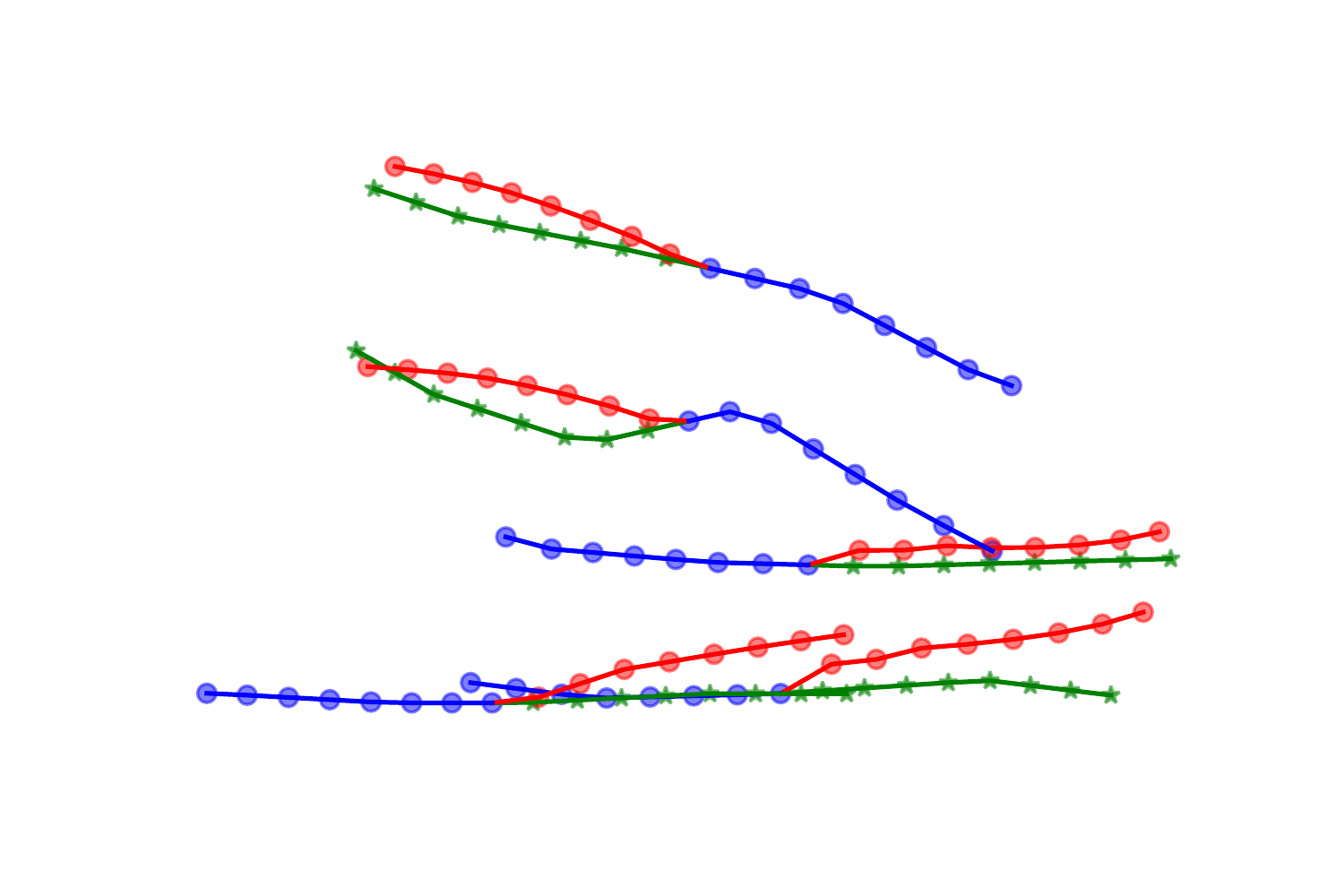} 
    \end{center}
\end{minipage}
&
\begin{minipage}{0.22\hsize}
    \begin{center}
        \includegraphics[clip, width=\hsize]{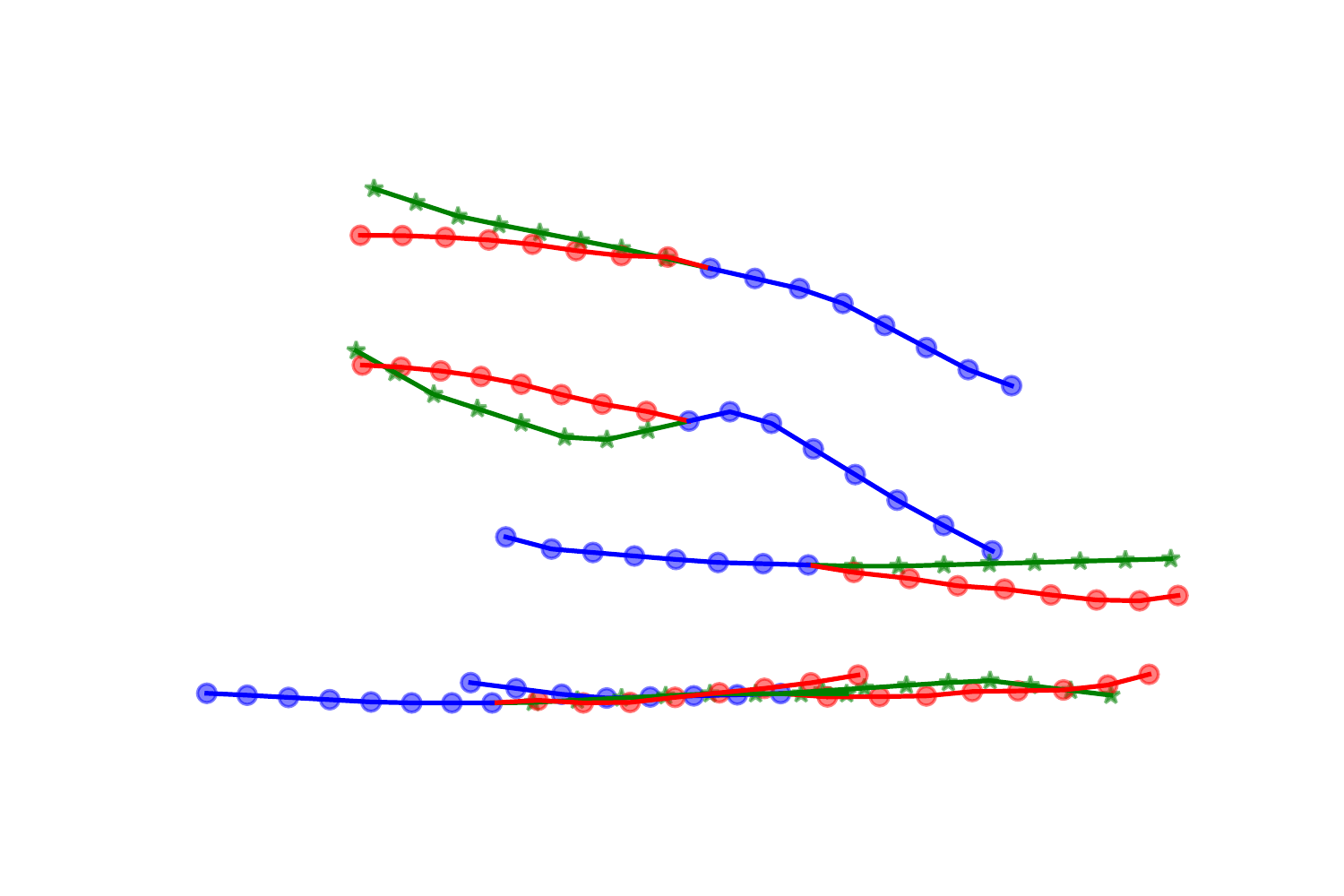}
    \end{center}
\end{minipage}
&
\begin{minipage}{0.22\hsize}
    \begin{center}
        \includegraphics[clip, width=\hsize]{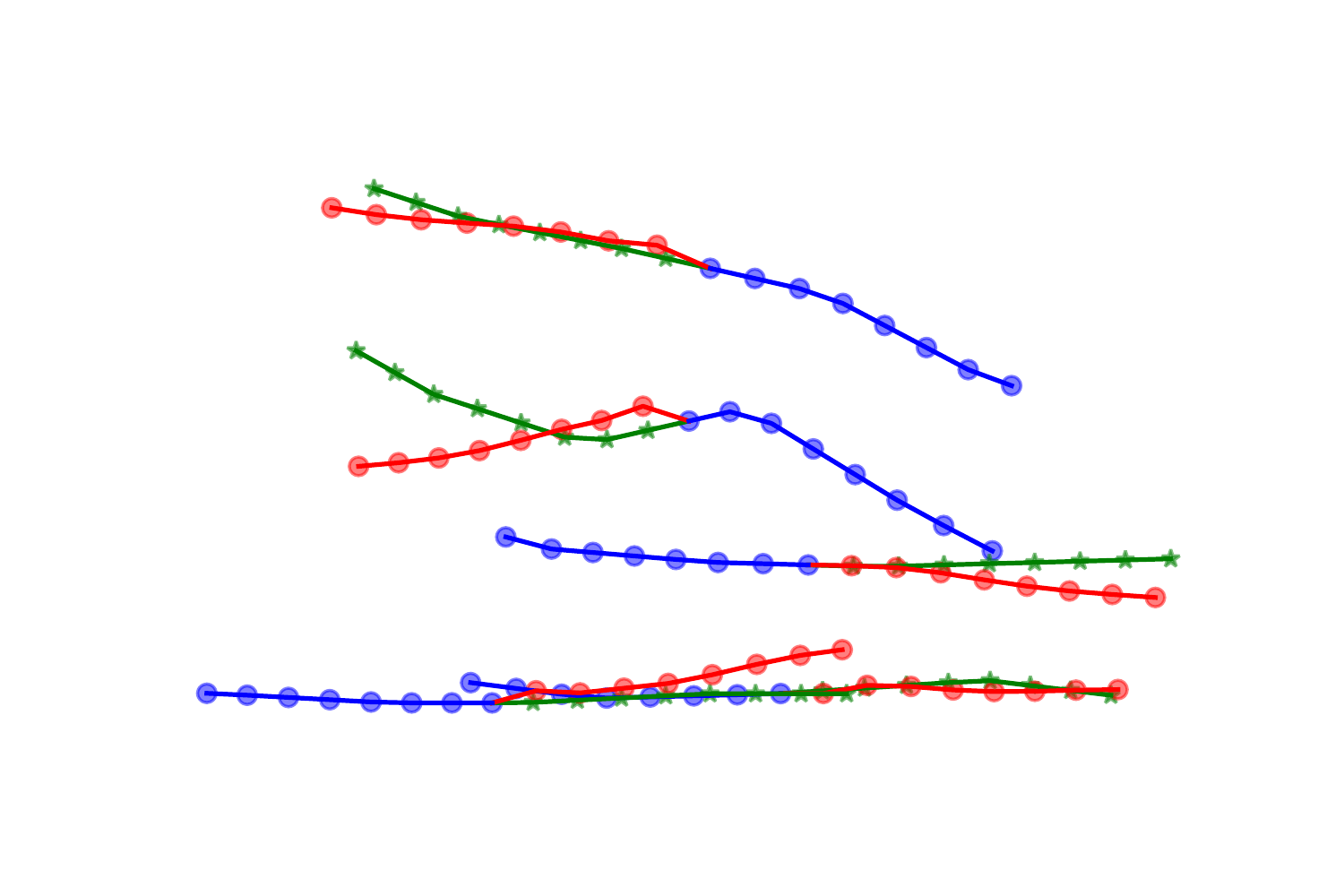}
    \end{center}
\end{minipage}
&
\begin{minipage}{0.22\hsize}
    \begin{center}
        \includegraphics[clip, width=\hsize]{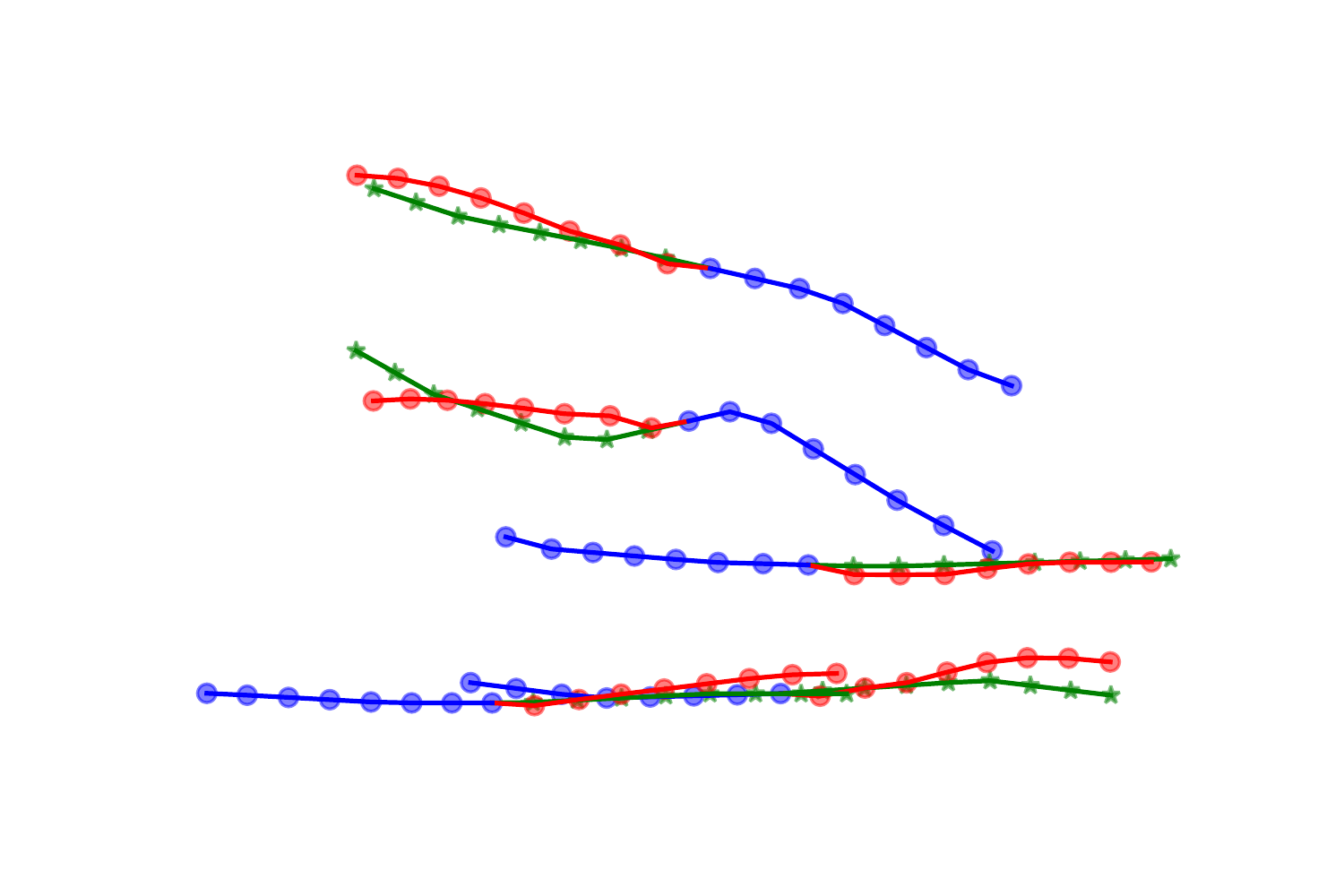}
    \end{center}
\end{minipage}
\vspace{-1.5em}
\\
\multicolumn{5}{c}{
\vspace{-2em}
        \includegraphics[clip, width=0.7\hsize]{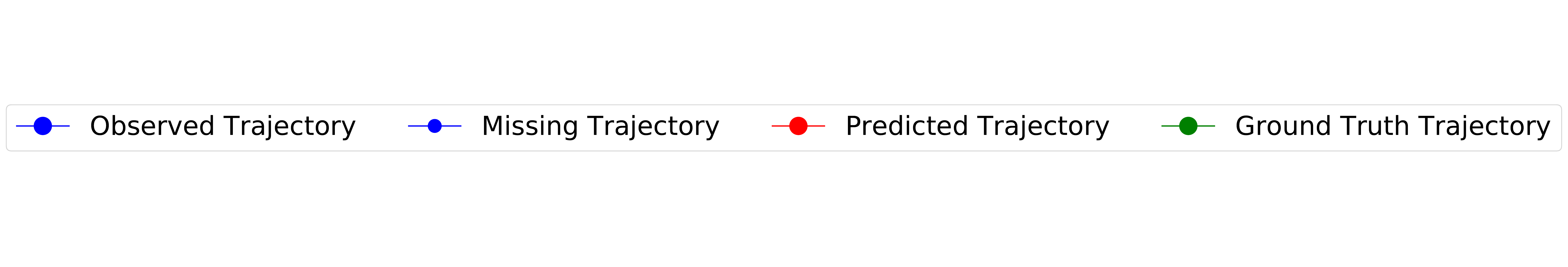}}
\end{tabular}
\caption{Qualitative comparison between our two-block RNN and the baseline models. (a) Trajectory prediction from observed trajectories with miss-detection and (b) trajectory prediction from the complete observed trajectories. 
}
\label{fig:8}
\end{center}
\end{figure*}

{\bf Qualitative evaluation.}
As a demonstration, we visualize the prediction results of our two-block RNN and those of the baseline methods in Fig. {\ref{fig:8}}. Here, we draw the average predicted trajectory of $100$ samples. In Fig. {\ref{fig:8}(a)}, we can observe that the prediction of our two-block RNN is more accurate than those of the baselines. Moreover, even when the complete trajectories are available (Fig. {\ref{fig:8}}(b)),  our two-block RNN can still predict the trajectories that are closer to the ground truth.

\section{Discussion}
In this section, we discuss some limitations. To begin with, we used the absolute coordinates, instead of relative coordinates. To compute the relative coordinates, we need the current position and the next position. Thus, it is not possible to compute both the previous and the next relative coordinates at one time step, when there is missing data. However, using absolute coordinates leads to a bad influence on  overall performance. It is especially the case that all models perform much more poorly than those trained with relative coordinates  for the task of trajectory prediction from complete data in all datasets ($64\%$ worse in $T_{pred}=8$ and $74\%$ worse in $T_{pred}=12$ on the ADE metric). If we could use relative coordinates for trajectory prediction with incomplete data, the performance might get better. In future work, we will work on this problem.

Next, we discuss other problems of trajectory prediction in real-world environments. We focus on predicting the future trajectory from incomplete observed trajectory due to miss-detection. Our model cannot handle the wrong detection, and tracking is out of focus, but if we can classify the detection or tracking result as valid or not, we can apply our model to these situations.

\section{CONCLUSION}
In this paper, we address the problem of trajectory prediction from incomplete observed trajectory because of miss-detection. We proposed a two-block RNN-based model, which takes advantage of the connection between the Bayesian filter and RNNs. Extensive experimental results on standard datasets, including ETH \cite{Pellegrini_2010_ECCV} and  UCY \cite{Leal_2014_CVPR}, show that our approach outperforms baselines that are commonly used. Since we do not use imputed data for training, our model does not affect the performance of trajectory prediction from the complete observed trajectory. 

\bibliographystyle{unsrt} 
\bibliography{refs} 

\clearpage
\begin{IEEEbiography}[{\includegraphics[width=1in,height=1.25in,clip,keepaspectratio]{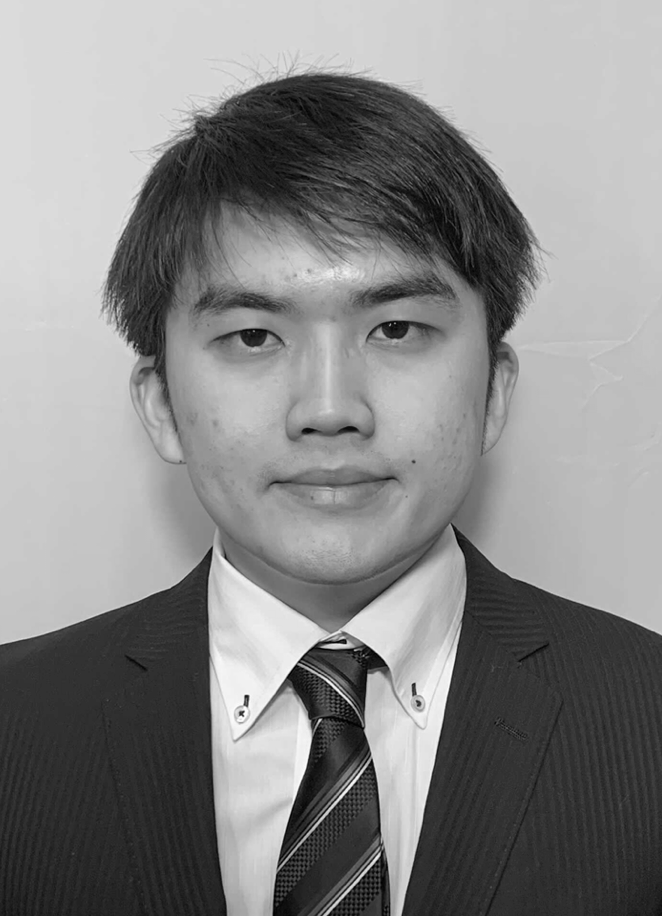}}]{Ryo Fujii} received the B.E. degree in information and computer science from Keio University, Japan, in 2020. He is currently pursuing the M.Sc.Eng. in information and computer science with Keio University, Japan.  His research interests include machine learning and computer vision.
\end{IEEEbiography}

\begin{IEEEbiography}[{\includegraphics[width=1in,height=1.25in,clip,keepaspectratio]{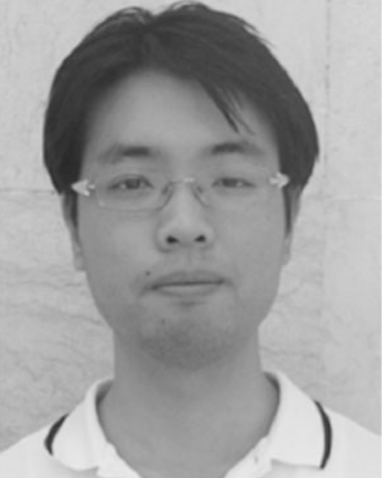}}]{Jayakorn Vongkulbhisal} received the B.Eng. degree in information and communication engineering from Chulalongkorn University, Bangkok, Thailand, in 2011; the M.Sc. degree in electrical and computer engineering from Carnegie Mellon University, Pittsburgh, PA, USA, in 2016; and the Dual-Degree Ph.D. degree in electrical and computer engineering from Carnegie Mellon University, Pittsburgh, PA, USA, and Instituto Superior Técnico, Lisbon, Portugal, in 2018. Since 2018, he has been a Research Scientist at IBM Research, Tokyo, Japan. His research interest includes optimization and machine learning in computer vision.
\end{IEEEbiography}

\begin{IEEEbiography}[{\includegraphics[width=1in,height=1.25in,clip,keepaspectratio]{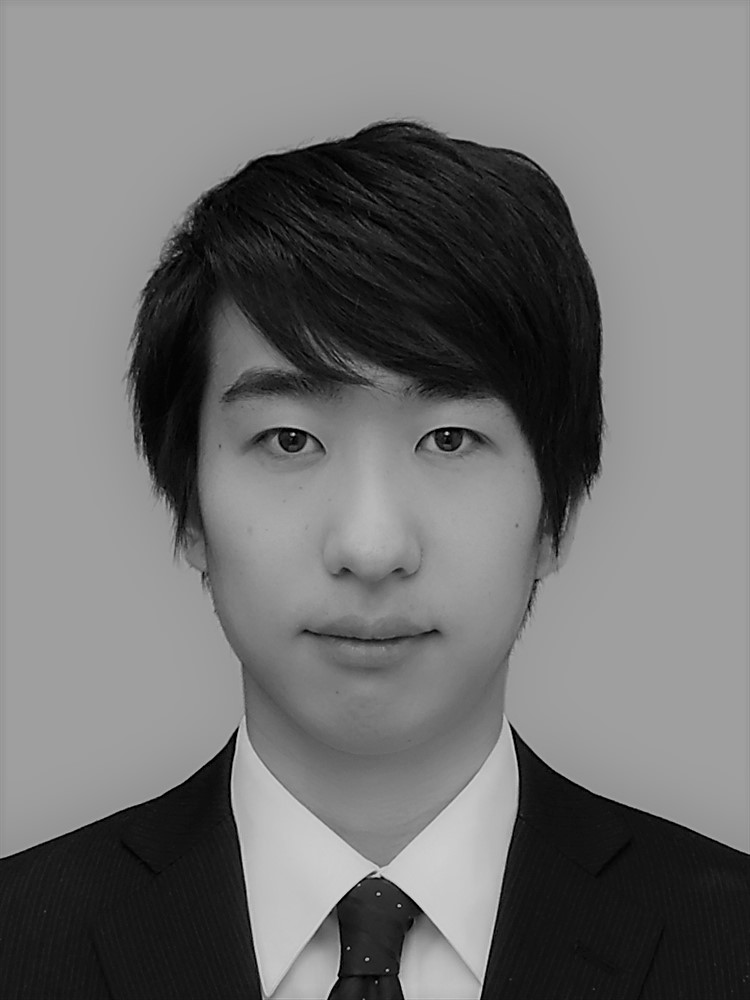}}]{Ryo Hachiuma} received his B.E. and M.Sc.Eng. degree in information and computer science from 
Keio University, Japan, in 2016 and 2017, respectively. He is currently pursuing the Ph.D. degree in science and technology with Keio University, Japan. His research interests include 3D object recognition, 3D object tracking, and simultaneous localization and mapping. 
\end{IEEEbiography}

\begin{IEEEbiography}[{\includegraphics[width=1in,height=1.25in,clip,keepaspectratio]{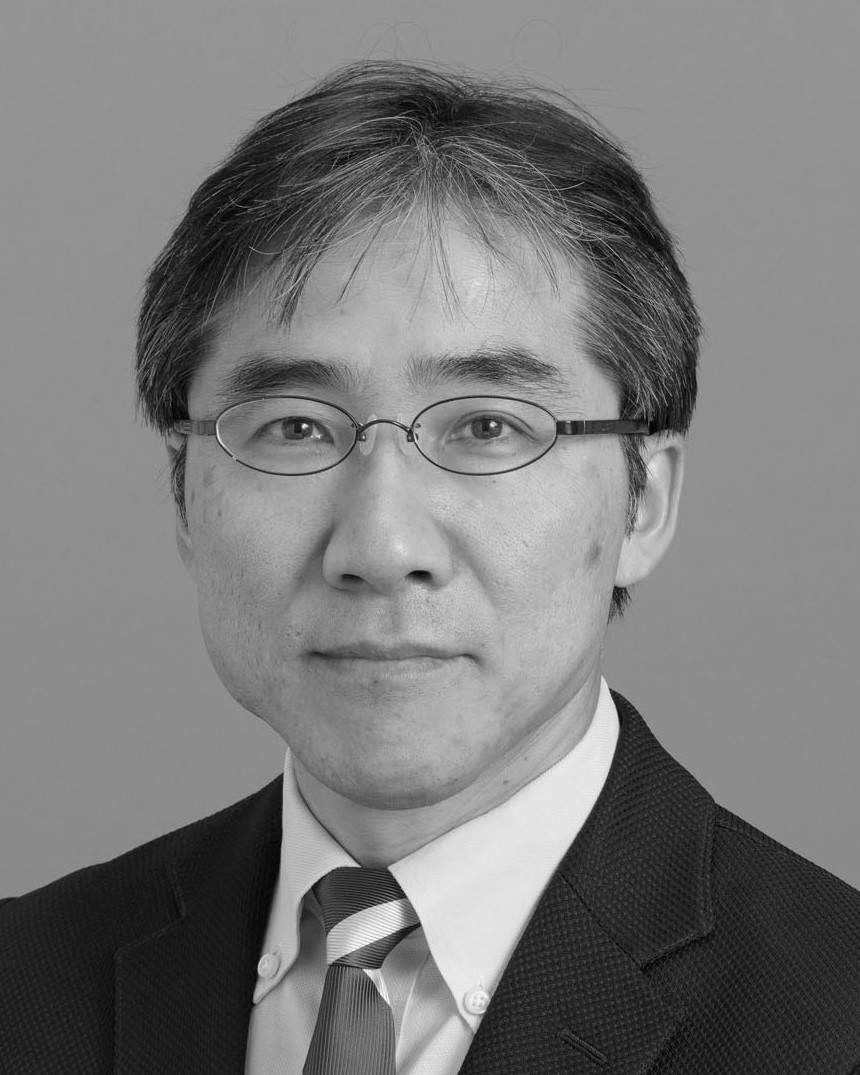}}]{Hideo Saito} received the Ph.D. degree in electrical engineering from Keio University, Japan, in 1992. Since 1992, he has been on the Faculty of Science and Technology, Keio University. From 1997 to 1999, he joined the Virtualized Reality Project at the Robotics Institute, Carnegie Mellon University, as a Visiting Researcher. Since 2006, he has been a Full Professor with the Department of Information and Computer Science, Keio University. 
\end{IEEEbiography}

\EOD

\end{document}